\documentclass[]{servicenow} 

\usepackage{amsmath,amsfonts,bm}

\def\eqref#1{equation~\ref{#1}}

\def\1{\bm{1}}

\DeclareMathAlphabet{\mathsfit}{\encodingdefault}{\sfdefault}{m}{sl}
\SetMathAlphabet{\mathsfit}{bold}{\encodingdefault}{\sfdefault}{bx}{n}

\usepackage{graphicx}

\usepackage{hyperref}
\usepackage{url}
\usepackage{booktabs}
\usepackage{multirow}
\usepackage{multicol}
\usepackage{titlesec}
\usepackage{subcaption}
\usepackage{cleveref}
\usepackage{wrapfig}
\usepackage[most]{tcolorbox}
\usepackage{algorithm}
\usepackage{algpseudocode}
\usepackage{amsmath}

\title{JEF-Hinter: Leveraging Offline Knowledge for Improving Web Agents Adaptation}

\author[1,2,3,4]{Hadi Nekoei}
\author[1,5]{Aman Jaiswal}
\author[1]{Patrice Bechard}
\author[1]{Oleh Shliazhko}
\author[1]{Orlando Marquez Ayala}
\author[2,3,4]{Mathieu Reymond}
\author[1]{Massimo Caccia}
\author[1,3]{Alexandre Drouin}
\author[2,3,6,7]{Sarath Chandar}
\author[1]{Alexandre Lacoste}

\affiliation[1]{ServiceNow AI Research}
\affiliation[2]{Chandar Research Lab}
\affiliation[3]{Mila -- Quebec AI Institute}
\affiliation[4]{Université de Montréal}
\affiliation[5]{Dalhousie University}
\affiliation[6]{Polytechnique Montréal}
\affiliation[7]{Canada CIFAR AI Chair}

\abstract{
Large language model (LLM) agents perform well in sequential decision-making tasks, but improving them on unfamiliar domains often requires costly online interactions or fine-tuning on large expert datasets. These strategies are impractical for closed-source models and expensive for open-source ones, with risks of catastrophic forgetting. Offline trajectories offer reusable knowledge, yet demonstration-based methods struggle because raw traces are long, noisy, and tied to specific tasks. We present \emph{Just-in-time Episodic Feedback Hinter (\MethodName{})}, an agentic system that distills offline traces into compact, context-aware hints. A zooming mechanism highlights decisive steps in long trajectories, capturing both strategies and pitfalls. Unlike prior methods, \MethodName{} leverages both successful and failed trajectories, extracting guidance even when only failure data is available, while supporting parallelized hint generation and benchmark-independent prompting. At inference, a retriever selects relevant hints for the current state, providing targeted guidance with transparency and traceability. Experiments on MiniWoB++, WorkArena-L1, and WebArena-Lite show that \MethodName{} consistently outperforms strong baselines, including human- and document-based hints.}

\newcommand{\MethodName}{{\normalsize\textsc{JEF-Hinter}}}

\begin{document}

\maketitle

\begin{figure}[ht]
    \centering
    \begin{subfigure}{0.32\textwidth}
        \centering
        \includegraphics[width=\linewidth]{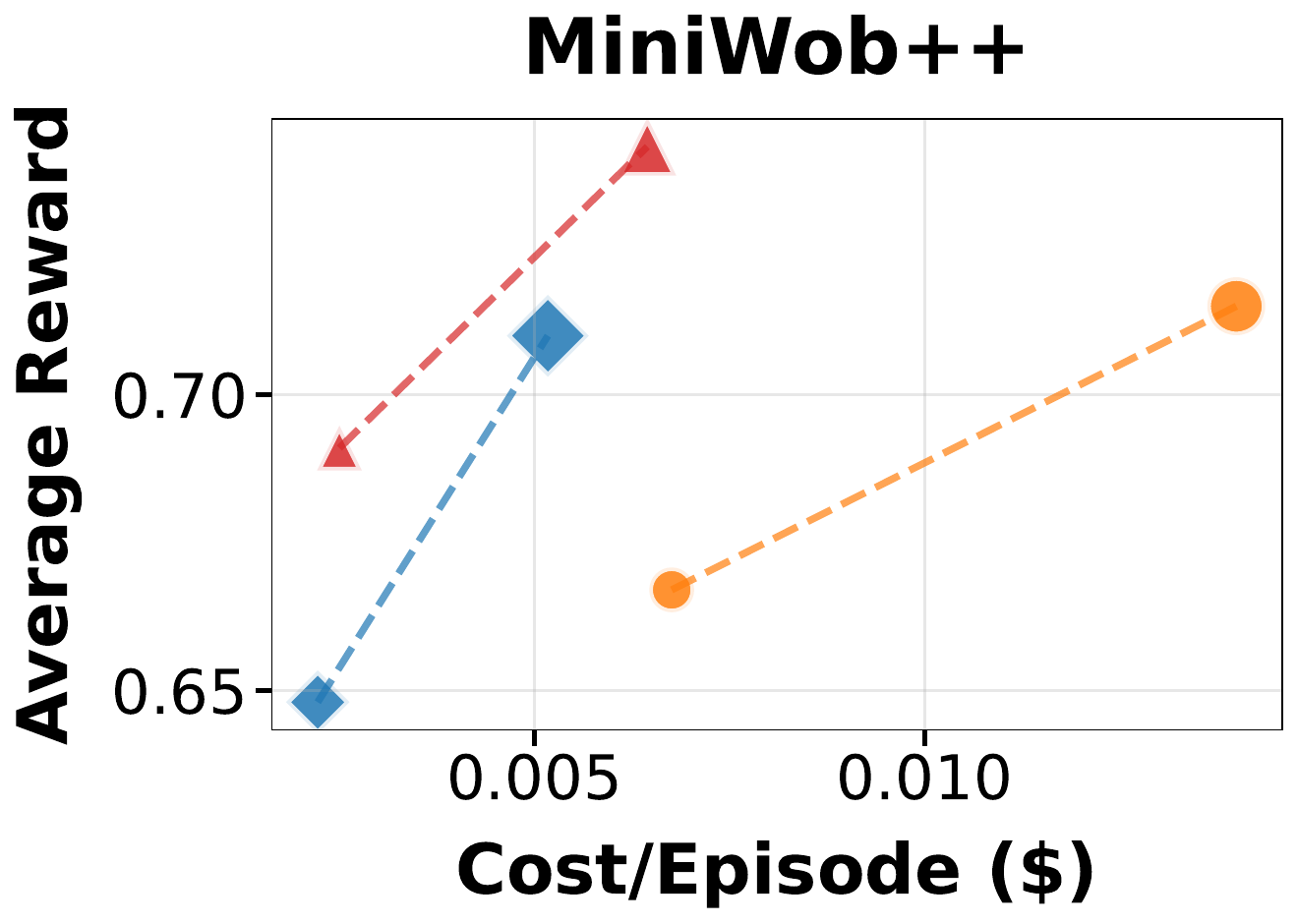}
    \end{subfigure}
    \hfill
    \begin{subfigure}{0.32\textwidth}
        \centering
        \includegraphics[width=\linewidth]{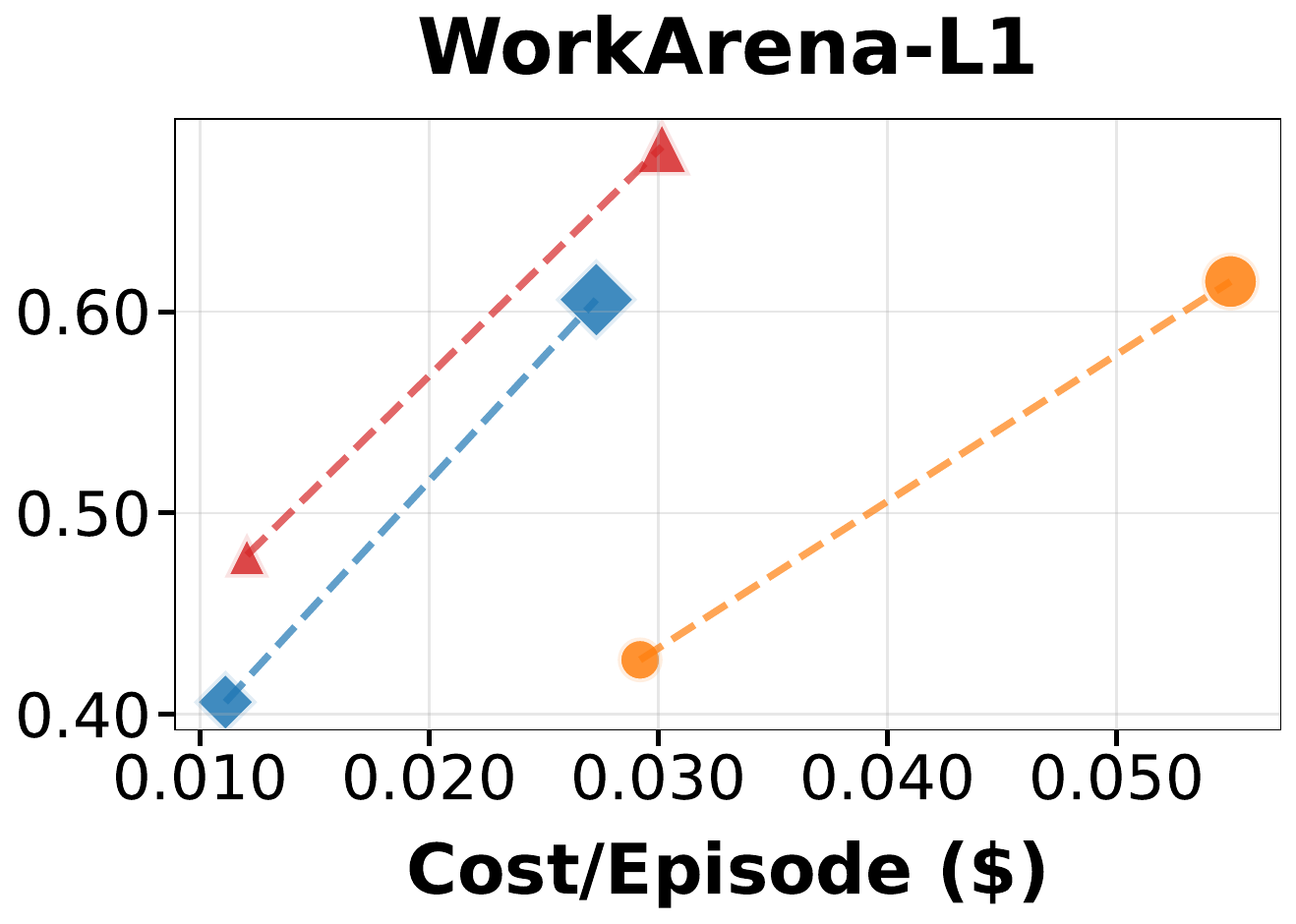}
    \end{subfigure}
    \hfill
    \begin{subfigure}{0.32\textwidth}
        \centering
        \includegraphics[width=\linewidth]{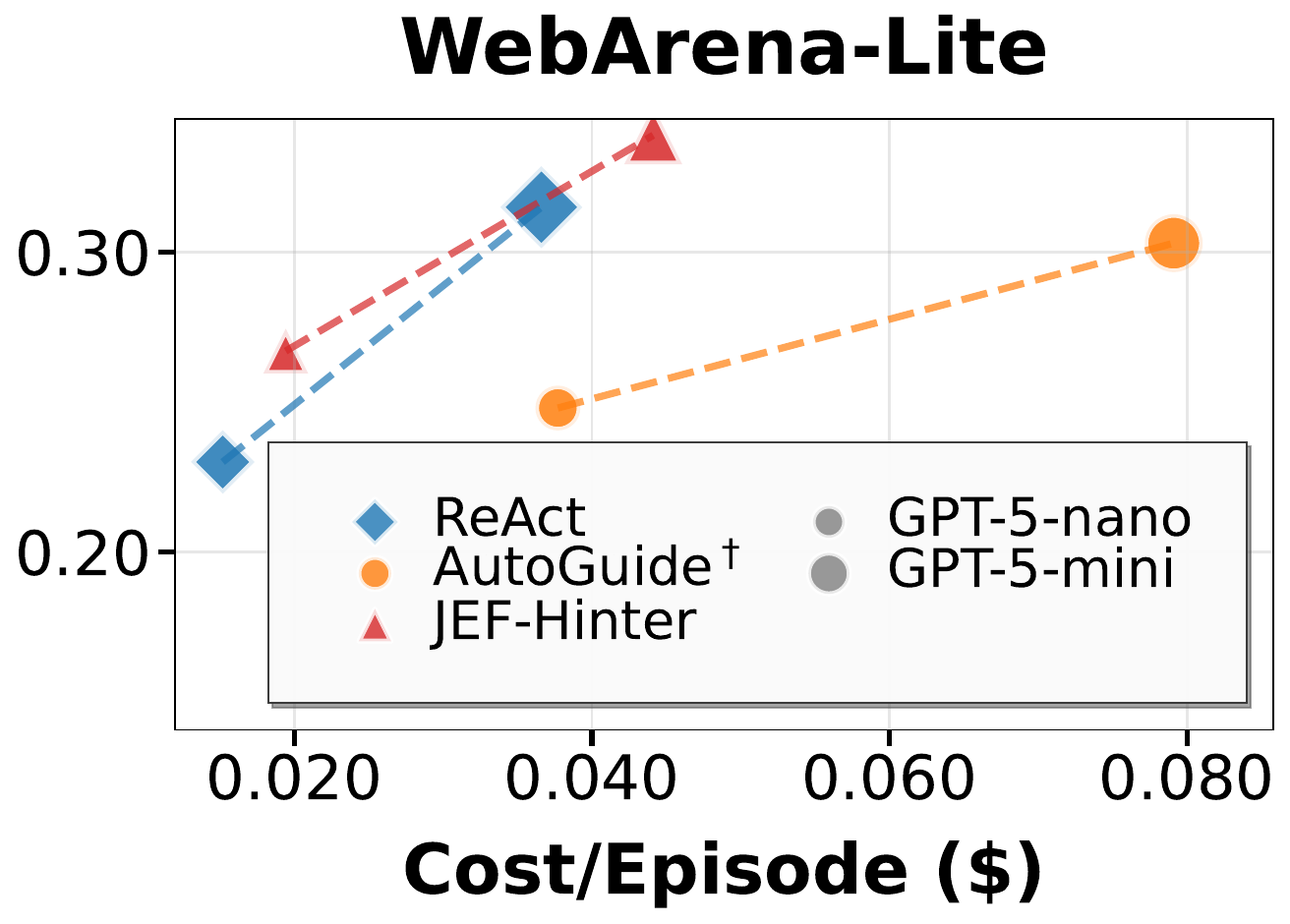}
    \end{subfigure}
    \caption{Average episodic reward versus test-time evaluation cost of \MethodName{} on MiniWoB++, WorkArena-L1, and WebArena-Lite, using \textbf{GPT-5-mini} as the Hinter model. Colors and markers denote different methods, while marker size reflects the base LLM model size. \MethodName{} achieves substantial gains over baselines, incurring only slightly higher cost than the original ReAct~\citep{yao2023react} agent while being far more efficient than Autoguide$^\dagger$~\citep{fu2024autoguide}.}

    \label{fig:2d_plots}
\end{figure}

\section{Introduction}

Large language model (LLM) agents have shown impressive abilities in sequential decision-making tasks such as web navigation and interactive environments. Yet their performance often deteriorates in unfamiliar domains due to incomplete domain knowledge and reasoning gaps.  
Offline resources offer an attractive opportunity. Trajectories from prior agents (both successful and failed), human demonstrations, and organizational documents all encode reusable decision patterns. In many practical deployments, these resources arise from tasks that recur frequently within the same domain or closely related domains, making the ability to systematically reuse past experience particularly valuable. Leveraging this knowledge is critical for closed-source models, which cannot be fine-tuned, and for large open-source models, where fine-tuning is costly and often risks catastrophic forgetting. Methods that can distill reusable knowledge from offline data provide a scalable way to improve state-of-the-art models without retraining or waiting for new releases.

Supervised fine-tuning on offline trajectories can appear to work, but off-policy bias means the learned policy cannot reliably execute even the training tasks end-to-end on its own, and it generalizes poorly to new tasks~\citep{ouyang2022training,yao2022webshop}. 
Reinforcement learning can be effective for web agents \citep{vattikonda2025train}, but its reliance on extensive online interactions is impractical at scale, and it cannot be applied to closed-source models.
Retrieval-augmented generation (RAG) methods, such as in-context demonstrations~\citep{lewis2020retrieval}, provide task-specific examples at inference, but raw trajectories are long, noisy, and tightly bound to their source tasks, limiting transfer. Recent work, such as AutoGuide~\citep{fu2024autoguide}, addresses part of this gap by distilling guidelines from offline trajectories, but it is limited to contrastive trace pairs and uses benchmark-specific prompting. These challenges motivate a more general and scalable framework for extracting and reusing offline knowledge.

We introduce \MethodName{}, an agentic system that distills offline traces into explicit, context-aware hints.
Instead of replaying full trajectories~\citep{shinn2023reflexion, fu2024autoguide}, \MethodName{} employs a \emph{zooming module} to identify critical decision points and a \emph{reflection step} to convert them into concise natural-language hints capturing both effective strategies and common pitfalls. Hints can be generated from single traces, pairwise contrasts, or multi-trace aggregation, ensuring coverage even when no successful run exists. Each hint is paired with a \emph{semantic key} for retrieval, enabling either fine-grained step-level guidance or efficient goal-conditioned retrieval at inference, preventing overload from irrelevant information~\citep{zhao2024expel} and complementing intra-task reflection mechanisms~\citep{shinn2023reflexion}. This offline-to-online pipeline produces a lightweight database of actionable hints that improves agent robustness and long-horizon performance without requiring model fine-tuning.

The algorithmic framework of \MethodName{} (comprising zooming, reflection, and retrieval) is domain-agnostic and can be applied to any sequential decision-making setting where offline trajectories exist. However, the \textit{hint databases} produced are domain-specific, as they capture patterns from particular environments and interfaces. We focus our evaluation on web navigation because long observations (HTML/AXTree) make zooming especially valuable, and benchmarks like MiniWoB++ and WorkArena support multiple task instantiations for studying within-environment hint transfer—the primary use case for organizations accumulating experience within workflows. 

Crucially, \MethodName{}'s benefits compound over time: as tasks recur, hints derived from past failures and successes reduce variance, prevent repeated errors, and stabilize agent behavior. Since \MethodName{} represents guidance as explicit hints linked to their source traces, it provides greater transparency and traceability than supervised fine-tuning or in-context RAG. Beyond effectiveness, \MethodName{} is cost-efficient at inference: hint generation is performed offline and amortized across executions, so test-time overhead is limited to lightweight retrieval and prompt augmentation. As shown in Figure~\ref{fig:2d_plots}, \MethodName{} achieves substantial gains while incurring only slightly higher cost than ReAct~\citep{yao2023react} and remaining more efficient than AutoGuide~\citep{fu2024autoguide}.

\textbf{Contributions:}
\begin{itemize}
    \item We introduce \emph{Just-in-time Episodic Feedback (\MethodName{})}, an agentic system that distills offline trajectories into explicit, context-aware hints. \MethodName{} features parallelized hint generation, intelligent zooming on critical steps, and flexible trace selection (single, pairwise, or multi-trace), leveraging both successful and failed runs.  
    \item We evaluate \MethodName{} across MiniWoB++, WorkArena-L1, and WebArena-Lite, where it consistently outperforms strong baselines. We further compare against documentation retrieval and human-authored hints, showing that automatically generated hints provide more scalable and broadly effective guidance.  
    \item We provide qualitative analyses that illustrate how \MethodName{} addresses common agent failure modes by steering actions toward the correct context and preventing repeated errors, thereby improving robustness and transparency.  
\end{itemize}

\section{Related Work}

LLMs have shown strong reasoning capabilities~\citep{wei2022chain}, resulting in LLM-based agents applied on a variety of real-world interactive tasks, including web navigation~\citep{nakano2021webgpt,wei2025webagentr,zhang2025webpilot}. However, performance on multiple web-focused benchmarks~\citep{yao2022webshop,deng2023mindweb,zhou2024webarena,koh2024visualwebarena,drouin2024workarena,boisvert2024workarena++} indicates that, as-is, LLMs still struggle with complex tasks requiring planning over long horizons. This gap has motivated several directions of work on improving LLM-based agents.

\textbf{Prompting and reflection.} A large body of work explores prompting strategies to elicit stronger reasoning and planning from LLMs. ReAct~\citep{yao2023react} interleaves reasoning steps with environment actions to structure trajectories. Building on this, Reflexion~\citep{shinn2023reflexion} introduces self-reflection over past trials to refine behavior, while ExpeL~\citep{zhao2024expel} mines offline Reflexion trajectories to extract reusable skills. Other approaches focus on explicit planning: AdaPlanner~\citep{sun2023adaplanner} iteratively adapts a plan to specific task instances, and AutoPlan~\citep{ouyang2023autoplan} instead optimizes for generalizable plans across instances. Methods such as Inner Monologue~\citep{huang2023inner} and Self-Refine~\citep{madaan2023self} further extend reflection by continuously revising intermediate reasoning. These approaches primarily improve agent behavior through online, in-context adaptation within or across episodes, rather than by distilling reusable guidance from offline experience.

\textbf{Search-based planning.} Beyond prompting, several works integrate symbolic search with LLM reasoning to better handle long-horizon tasks. Tree-of-Thoughts~\citep{yao2023tree}, Language Agent Tree Search~\citep{zhou2024language}, and their variants~\citep{putta2024agent,koh2025tree} explore branching reasoning paths and dynamically select among them, improving robustness on tasks where a single-line chain-of-thought often fails. 
These methods are complementary to ours: search improves robustness through additional test-time computation, whereas our approach improves reliability by amortizing experience from offline trajectories.

\textbf{Offline data and hinting.}
Orthogonal to online prompting and search, another line of work focuses on extracting reusable guidance from offline data.
Retrieval-augmented generation (RAG) approaches~\citep{lewis2020retrieval} have been adapted for agents by retrieving demonstrations or examples~\citep{yao2023react}, but raw trajectories are often long, noisy, and tightly coupled to their source tasks, which limits their transferability.
AutoGuide~\citep{fu2024autoguide} addresses part of this challenge by distilling abstract guidelines from contrastive trajectory pairs, demonstrating that such distilled guidance can outperform raw demonstrations.
More recent work explores structured reuse of offline experience in different forms.
Agent Workflow Memory (AWM)~\citep{wang2024agentworkflowmemory} induces reusable workflows from successful trajectories, allowing agents to accumulate and replay subroutines across tasks.
AutoManual~\citep{chen2024automanual} similarly constructs reusable manuals, but relies on online interaction and iterative rule refinement.
Concurrent memory-based approaches such as Agentic Context Engineering (ACE)~\citep{zhang2026agenticcontextengineeringevolving} and ReasoningBank~\citep{ouyang2025reasoningbankscalingagentselfevolving} further show that storing and retrieving structured reasoning or procedural memories can improve agent performance over repeated executions.

While these methods demonstrate the promise of reusing past experience, they typically operate at the level of workflows, procedures, or evolving memory structures, and often assume access to successful executions or continual interaction to refine stored knowledge.
In contrast, \MethodName{} focuses on distilling \emph{lightweight, context-aware natural language hints} from heterogeneous offline data, including both successful and failed trajectories, without requiring contrastive pairs or online refinement.
Rather than replaying entire workflows or growing persistent memory structures, \MethodName{} isolates critical decision points via a zooming mechanism and converts them into concise guidance that captures both effective strategies and common pitfalls.
These hints are indexed by semantic keys and retrieved just-in-time, enabling fine-grained or goal-conditioned guidance with bounded inference-time overhead.

\section{Just-in-time Episodic Feedback Hinting}\label{sec:method}

\begin{figure}[ht]
    \centering
    \includegraphics[width=0.8\textwidth]{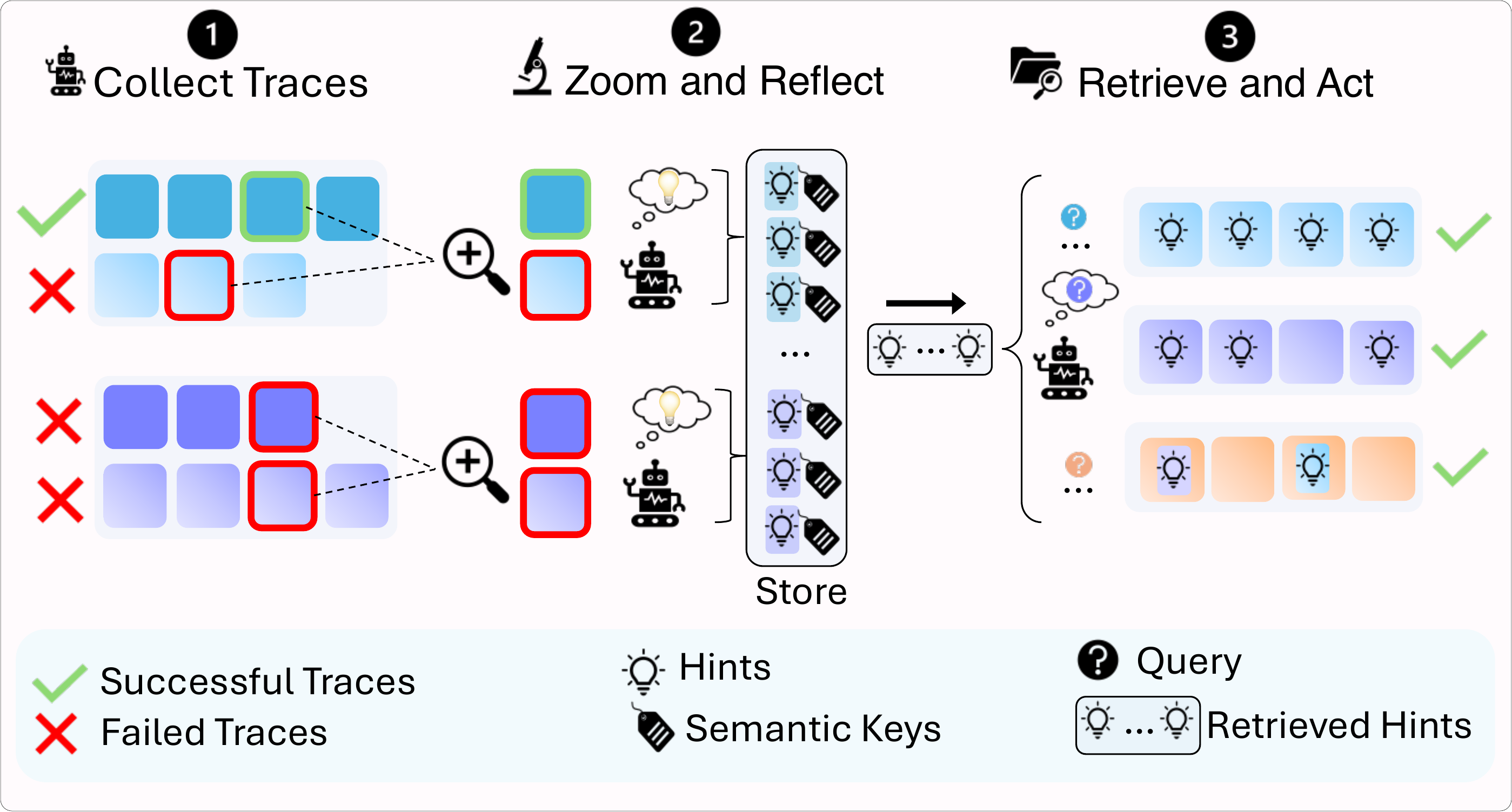}
    \caption{
    Overview of the \MethodName{}. 
    \textbf{(1) Collect Traces:} \MethodName{} operates over heterogeneous offline trajectories, including both successful (green) and failed (red) runs, allowing the system to capture not only effective behaviors but also common pitfalls. 
    \textbf{(2) Zoom and Reflect:} A zooming module selects critical steps within each trace, and the hinter reflects on these segments to distill them into concise, reusable natural language hints. Each hint is paired with a semantic key summarizing its context and stored for retrieval. 
    \textbf{(3) Retrieve and Act:} At inference time, the agent generates a query (goal- or context-conditioned) which is matched against the database of semantic keys. The most relevant hints are retrieved and injected into the agent's context, guiding its actions. This process unifies knowledge distillation, reflection, and retrieval, supporting both in-task reliability and out-of-task generalization.
    }
\end{figure}

Large language model (LLM) agents often struggle to generalize across tasks when relying solely on their base policy \(\pi\). Direct fine-tuning can be costly, unstable, or even impossible for closed-source models. To address this, we propose to improve \(\pi\) by supplying it with targeted, reusable guidance extracted from offline experience. At the center of our approach is the \emph{Hinter} \(\mathcal{H}\), itself an LLM, transforms trajectories and documents into explicit natural-language hints. Since hint generation is performed offline, \(\mathcal{H}\) can be significantly larger and more capable than the base agent, yet the resulting hints remain lightweight at inference. We instantiate this method as \MethodName{}, which systematically augments the LLM base policy with retrieved hints to enhance decision-making without any fine-tuning.

\subsection{Data Collection}

Unlike prior work such as AutoGuide~\citep{fu2024autoguide}, which extracts guidance only from contrastive trajectory pairs, \MethodName{} can operate over a broader range of offline signals. Given a dataset of trajectories \(\{\tau_1,\ldots,\tau_N\}\), it flexibly selects evidence for hint generation. The trajectories may come from the base policy \(\pi\), which yields hints tailored to its strengths and weaknesses, but they can also originate from other agents or human demonstrations. We support three complementary modes:

\begin{enumerate}\itemsep 0pt
    \item \textit{Single-trace analysis.} Generate hints from a single trajectory \(\tau\), highlighting effective decisions in successful segments and exposing pitfalls in failed ones.
    \item \textit{Pairwise analysis.} Contrast two trajectories \((\tau^+,\tau^-)\) where the total reward assigned to \(\tau^+\) is greater than the reward assigned to \(\tau^-\), and identify the key divergences that explain the performance gap. If no such pair is available, we also allow equal-reward or $(\text{fail},\text{fail})$ and $(\text{success},\text{success})$ pairs.
    \item \textit{Multi-trace analysis.} Combine a set of trajectories \(\{\tau^i\}_{i\in S}\) to surface patterns that are robust across instances and transferable across tasks.
\end{enumerate}

\subsection{Hint Generation: Zoom \& Reflect}

A trajectory provides four types of signals: observations \(x\) such as screenshots, HTML (or AxTree); reasoning tokens \(z\) that record intermediate thoughts; actions \(a\) that alter the environment; and rewards \(r\) that measure progress. The initial observation \(x_0\) also contains the goal \(g\). We combine these signals to form the prompt $(P)$ given to the Hinter. The simplest option is the full prompt $P_\tau^{\text{full}} = \{x,z,a,r\}_{1:T}$, which passes the entire trajectory as context. Long-horizon tasks quickly make this representation unwieldy. To address this, we introduce a Zooming LLM module that selects critical steps \(t^\ast\) and extracts a compact prompt: $$P_\tau^{\text{zoom}} = \{z,a,r\}_{1:T} \;\cup\; \{x\}_{t^\ast:t^\ast+\Delta}.$$
This keeps the full sequence of reasoning, actions, and rewards, while restricting observations to the decisive windows. The parameter $\Delta$ specifies the length of the observation window appended after $t^\ast$, determining how much context is retained. 
The zooming mechanism is particularly crucial for web-based tasks: a single step's HTML or AXTree observation can exceed 10K tokens, and long-horizon tasks may span 20+ steps, making full trajectory processing prohibitively expensive. By focusing on critical decision points, we reduce context by 80-95\% while improving hint quality, as the hinter receives high-signal context rather than being overwhelmed by repetitive page structures.
Critical steps correspond to points where the agent makes an important choice, repeats a common mistake, executes a successful strategy, interacts with a key element, handles a timing dependency, or reaches a definitive outcome. 
For instance, in a web form task, repeatedly clicking the wrong navigation bar is flagged as a critical step, while in a multi-select list, the decisive step is holding Ctrl/Cmd to select multiple items.
Appendix~\ref{appendix:zooming_prompt} details the step-selection procedure.


Next, to support retrieval, we generate a \emph{semantic key} summarizing the trajectory prefix. Given $\tau_{:t}$, the summarizer $\mathcal{S}$ outputs a short natural-language context $c_t = \mathcal{S}(\tau_{:t})$. This key anchors hint generation during training and enables efficient lookup at inference.

Finally, given a context $c_t$ and a prompt $P_\tau$, the Hinter produces a hint
$h = H(c_t, P_\tau)$,
which captures either a beneficial action or a common error to avoid. We collect all hints in a database $\mathcal{D}_\mathcal{H} = \{(c_t, h)\}$, linking each hint to the semantic key from which it was derived (see Appendix~\ref{appendix:algorithm} for pseudocode).

\subsection{Retrieve \& Act}

We explore two complementary strategies for retrieving and applying hints during inference.

\paragraph{Contextual retrieval with step-level hints.}  
At each time step \(t\), the summarizer produces a context \(c_t = \mathcal{S}(\tau_{:t})\).  
The retrieval LLM module $\rho$ then selects the top \(k\) hints most relevant to that context,
$\{h_t^{1},\ldots,h_t^{k}\}=\rho(c_t,\mathcal{D}_{\mathcal{H}})$, and the policy conditions its next action on both the trajectory prefix and the retrieved hints, $a_t \sim \pi\!\big(x_{0:t}, \{h_t^{1},\ldots,h_t^{k}\}\big)$. This approach provides fine-grained, context-specific guidance, but it is computationally costly since it requires one model call to establish the context and retrieve hints and another to generate the action.

\paragraph{Goal-conditioned retrieval with episode-level hints.}  
A more efficient strategy retrieves hints once at the start of an episode, using the goal \(g\) as the retrieval context: $\{h^{1},\ldots,h^{k}\}=\rho(g,\mathcal{D}_{\mathcal{H}})$. The policy then acts while simultaneously selecting a relevant hint from this fixed set, $(a_t,h_t) \sim \pi\!\big(x_{0:t}, \{h^{1},\ldots,h^{k}\}\big)$. 
This method avoids repeated retrieval calls and reduces inference cost, while still maintaining sufficient contextual relevance.

\paragraph{Source tasks for retrieval}  

The choice of source tasks also determines how well hints generalize.
In-task retrieval draws hints from the same task but with different goals\footnote{Benchmarks like MiniWoB++ and WorkArena support multiple seeds per task. We refer to a specific instance of a task as a goal.}, which strengthens reliability within a domain.  
Cross-task retrieval excludes the source task altogether and forces the agent to transfer knowledge from other tasks.  
Hybrid retrieval mixes both approaches with adjustable weighting, striking a balance between reliability and transfer.  
Because hints capture abstract decision patterns rather than raw demonstrations, they remain effective across goals and tasks under both settings.

The zooming module, summarizer~$\mathcal{S}$, hinter~$\mathcal{H}$, retriever~$\rho$, and base policy~$\pi$ are all LLM-based components. In contrast, the hint database~$\mathcal{D}_{\mathcal{H}}$, its indexing and storage, and the embedding-based matching used for retrieval are lightweight non-LLM operations responsible for orchestration and lookup.

\section{Experimental Setup}\label{sec:exp_setup}

\paragraph{Benchmarks}
We evaluate on three widely used benchmarks that span increasing levels of complexity: 
MiniWoB++~\citep{liu2018reinforcement}, a suite of synthetic single-page UI tasks; 
WorkArena-L1~\citep{drouin2024workarena}, a benchmark of enterprise knowledge-work tasks involving multi-step form filling and navigation; 
and WebArena~\citep{zhou2024webarena}, a realistic environment of multi-domain web tasks requiring long-horizon reasoning. 
Together, these benchmarks test both short-horizon precision and long-horizon generalization.

\begin{figure}[ht]
    \centering
    \includegraphics[width=\textwidth]{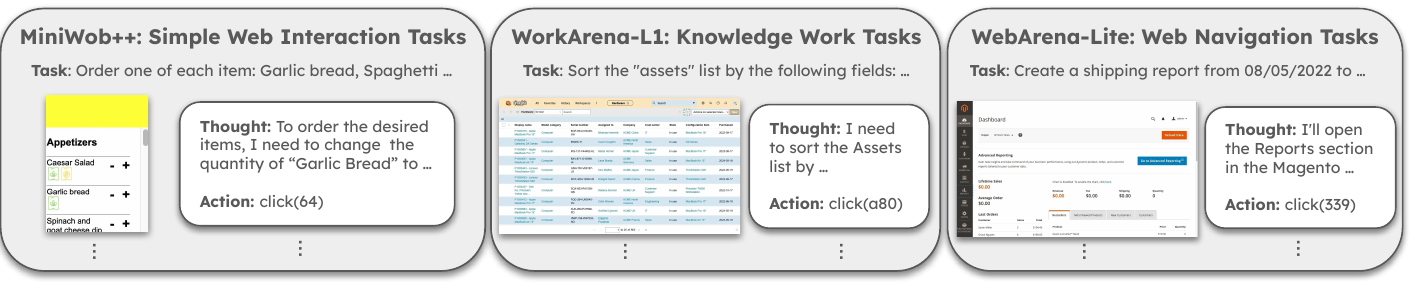}
    \caption{Web browsing benchmarks considered in our work: 
    MiniWob++~\citep{liu2018reinforcement}, 
    WorkArena-L1~\citep{drouin2024workarena}, 
    and WebArena-Lite~\citep{zhou2024webarena, liu2025visualagentbench}.}
\end{figure}

\paragraph{Observation and action spaces}
To improve speed and efficiency, we work with the accessibility tree (AXTree). This reduces the size of the input by about 10x compared to the HTML DOM trees. Exceptionally, on MiniWoB++, we work directly with the DOM since it is small enough and contains more of the relevant information.
The action space across all environments consists of high-level UI primitives such as \texttt{click(node)}, \texttt{fill(node, text)}, \texttt{select(node, option)}, \texttt{scroll(node)}, and \texttt{hover(node)} as provided by BrowserGym.
This abstraction enables consistent evaluation across benchmarks with differing interfaces.

\paragraph{Baselines}
All methods build on the ReAct agent framework~\citep{yao2023react}, which combines chain-of-thought reasoning with environment interaction. 
We compare against: (i) \textbf{ReAct} without offline hinting, (ii) Our implementation of \textbf{AutoGuide}~\citep{fu2024autoguide}, which augments ReAct with offline guideline extraction from contrastive trajectory pairs. We call this agent AutoGuide$^\dagger$.
\footnotetext{$^\dagger$Since no public implementation of AutoGuide was available, we re-implemented it within our ReAct framework for consistency and comparability.} 
In addition, we evaluate two variants of our agent: \textbf{\MethodName{}} (w/o zoom), our basic implementation that takes the full trajectory as input and distills offline trajectories into natural-language hints (for WorkArena-L1 and WebArena-Lite, we drop AxTrees to fit within the hinter model’s context), and \textbf{\MethodName}, which further includes zooming on critical steps.

\paragraph{Offline datasets} 
We construct offline datasets using the \textsc{AgentLab} framework~\citep{drouin2024workarena,dechezelles2025browsergymecosystemwebagent}.
For MiniWoB++, we collect trajectories by running a ReAct agent on 5 held-out goals per task, and for WorkArena-L1, we collect trajectories on 10 held-out goals per task. 
For WebArena, we use WebArena-Lite \cite{liu2025visualagentbench} for parallel trace collection. 
In all benchmarks, we retain both successful and failed trajectories so that hint extraction can cover both positive decision points and common pitfalls. 
In contrast, AutoGuide~\citep{fu2024autoguide} requires pairs of successful and failed traces and therefore only produces hints when both are available. 
To study the impact of dataset quality, we additionally construct augmented datasets by including traces from GPT-5, ensuring at least one successful trace per task. 
If no successful trace exists, even after augmentation, AutoGuide produces no hint for that task, whereas \MethodName{} can still generate useful hints from failed trajectories alone.

\paragraph{Evaluation protocol and generalization scope}
We evaluate within-environment generalization across different goals and tasks, reflecting how organizations accumulate experience from recurring workflows to improve agent reliability over time. We focus on web navigation because these benchmarks provide standardized protocols and represent high-impact applications where long-horizon tasks with large observation spaces (HTML/AXTree) make our zooming mechanism particularly valuable.
We evaluate generalization under two complementary settings: \emph{In-task generalization:} The agent retrieves hints only from the same source task, but from different goals than those used in evaluation, measuring how well hints transfer within a task across different environment initializations. \emph{Out-of-task generalization:} We exclude the source task entirely from the hint database. At inference time, the agent must rely on hints retrieved from other tasks using the LLM retriever or embedding vector matching. This tests whether hints can transfer effectively to unseen tasks with different structures—for instance, hints from filtering tasks helping with sorting tasks when they share underlying interaction patterns (list manipulation, form controls).
The primary evaluation metric is average task success rate, reported separately for in-task and out-of-task settings. We also provide qualitative analysis of hints to illustrate their interpretability and usefulness.

\section{Empirical Study}

We present results through research questions examining the effectiveness, generalization, and design decisions of \MethodName{}.

\begin{figure}[ht]
    \centering
    \begin{subfigure}{0.28\textwidth}
        \centering
        \includegraphics[width=\linewidth]{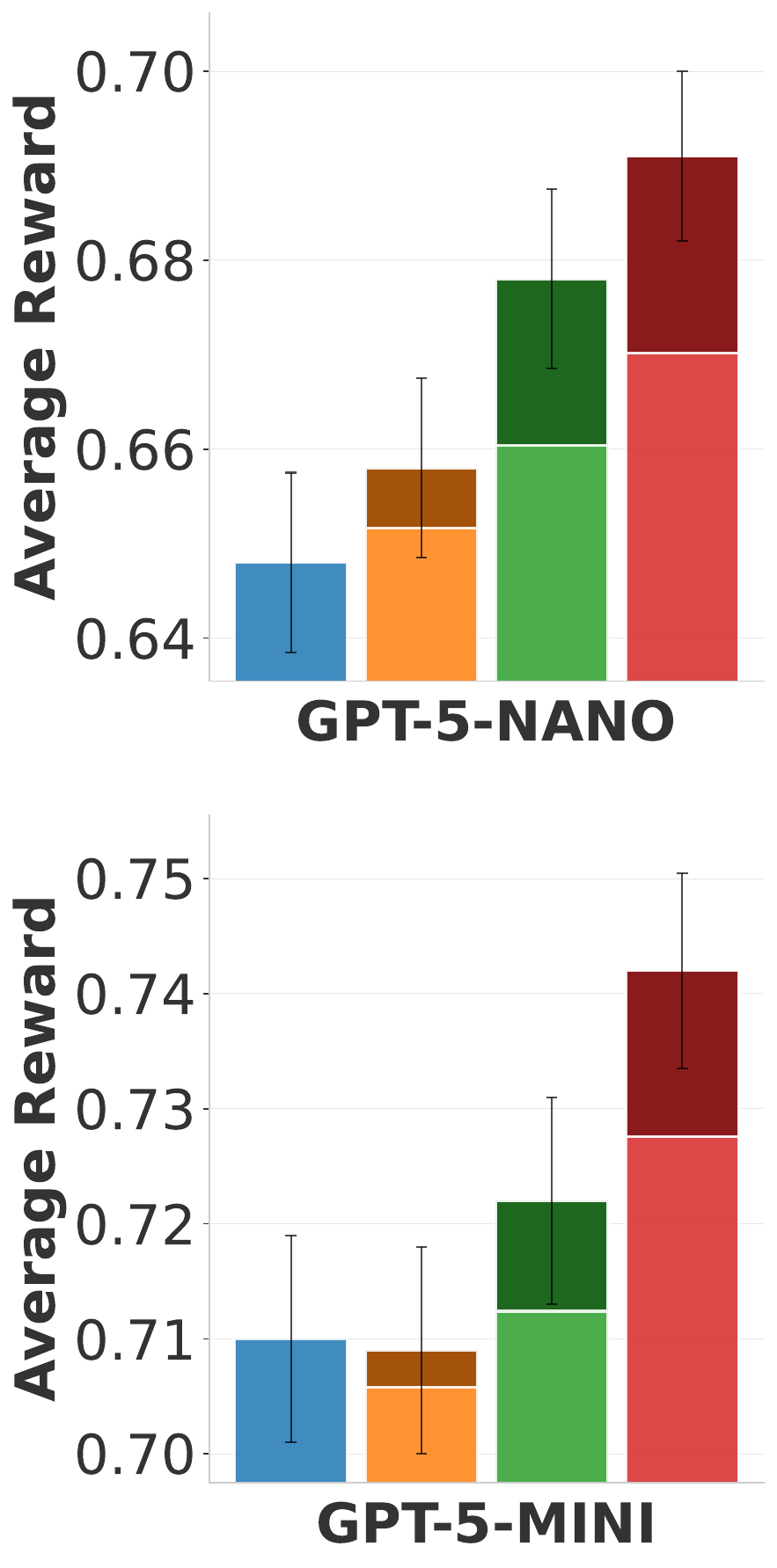}
        \caption{Miniwob++}
    \end{subfigure}
    \hfill
    \begin{subfigure}{0.28\textwidth}
        \centering
        \includegraphics[width=\linewidth]{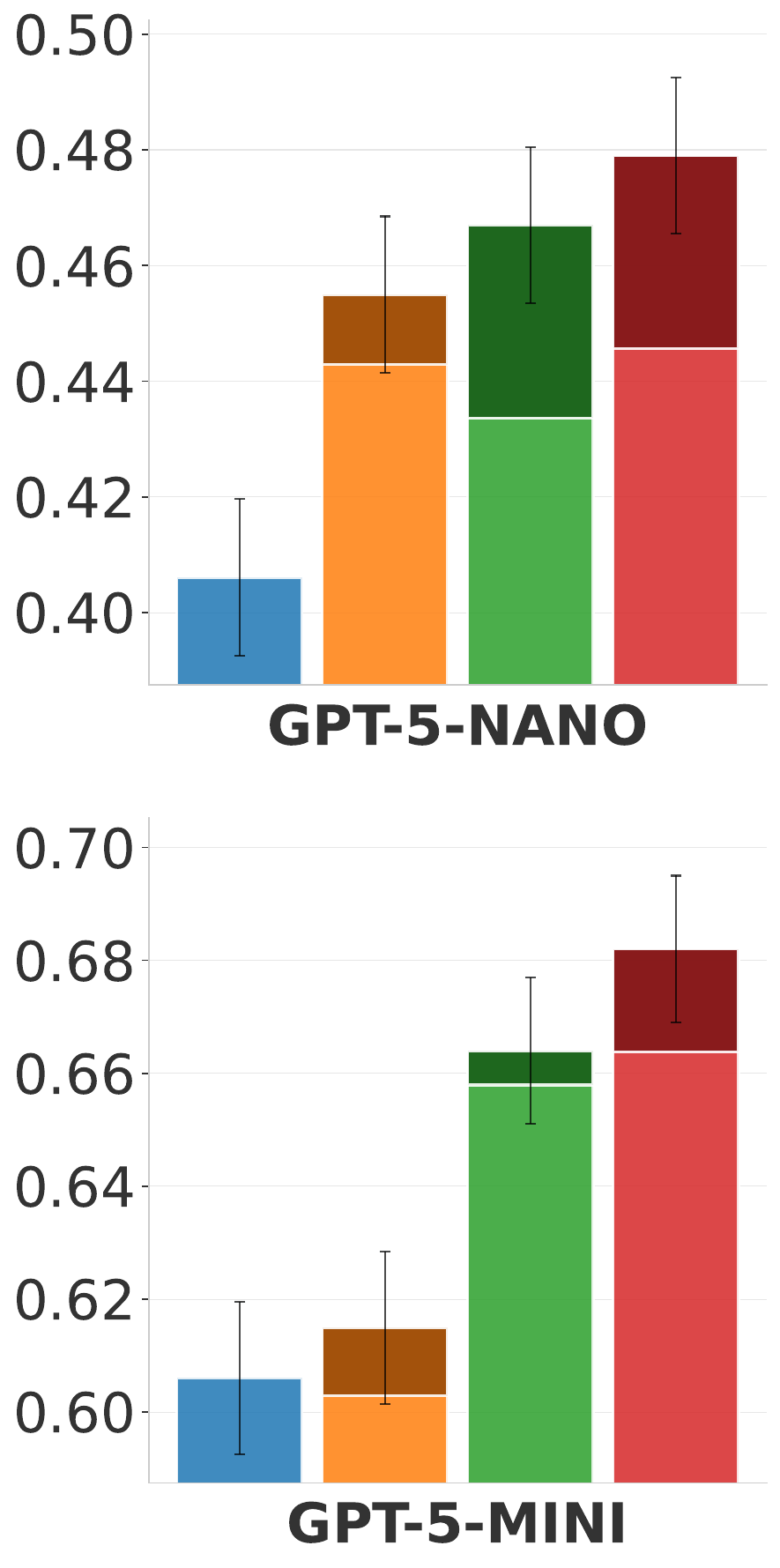}
        \caption{WorkArena-L1}
    \end{subfigure}
    \hfill
    \begin{subfigure}{0.28\textwidth}
        \centering
        \includegraphics[width=\linewidth]{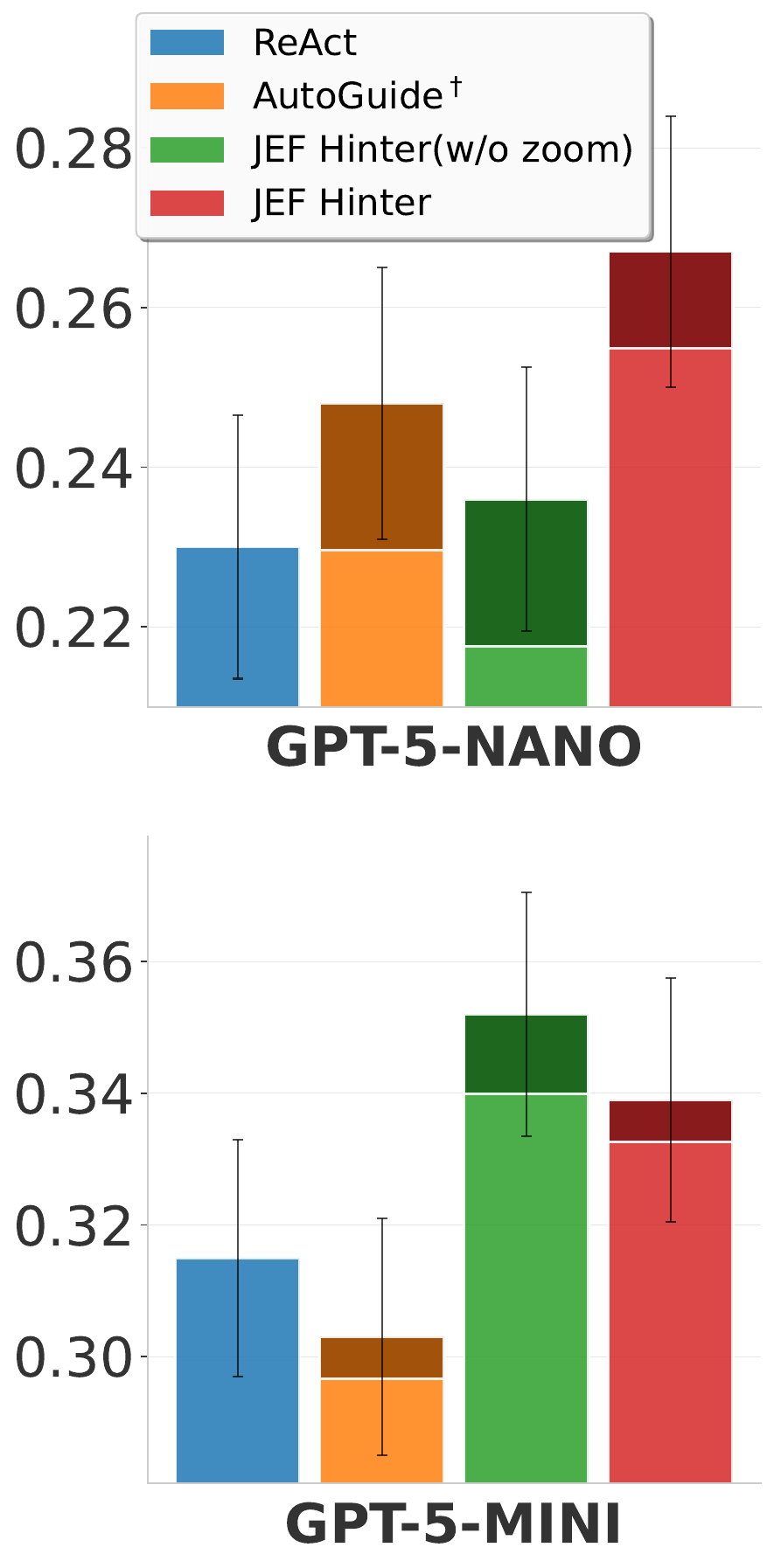}
        \caption{WebArena-Lite}
    \end{subfigure}
    \caption{Average reward comparison across MiniWoB++, WorkArena-L1, and WebArena-Lite using two base models with \textbf{GPT-5-mini} as the Hinter model. \MethodName\; and \MethodName(w/o zoom)\; consistently outperform all baselines across most tasks, highlighting the effectiveness of our approach. Shaded regions denote tasks where the base ReAct agent failed entirely, highlighting \MethodName{}’s ability to extract useful hints even from failure-only trajectories.}

    \label{fig:main_results}
\end{figure}

\subsection{Does \MethodName{} improve overall performance compared to baselines?}  

To address this question, we compare ReAct, AutoGuide, and \MethodName{} across three benchmarks: MiniWoB++, WorkArena-L1, and WebArena-Lite. As shown in \cref{fig:main_results}, three key findings emerge:
\paragraph{Generally, hints provide effective guidance to agents.}
Both AutoGuide and \MethodName{} consistently outperform ReAct across all benchmarks and base models. This confirms that offline hints provide meaningful guidance, steering the agent away from common pitfalls and toward more successful strategies. Moreover, since GPT-5-mini is used as the hinter model, the gains observed when the base model itself is GPT-5-mini highlight that \MethodName{} enables effective \emph{self-improvement}, demonstrating that a model can refine its own decision-making by reflecting on past traces.

\par\medskip
\begin{wraptable}{r}{0.45\columnwidth}
\centering
\small
\caption{Ablation of full-trace (FT) vs.\ zoomed multi-trace hinting.}
\label{tab:zoom_ablation}
\begin{tabular}{lcc}
\toprule
\textbf{Method} & \textbf{MiniWoB++} & \textbf{WorkArena-L1} \\
\midrule
ReAct               & 0.715 & 0.661 \\
\MethodName{} (FT)  & 0.718 & 0.715 \\
\MethodName{}       & 0.739 & 0.770 \\
\bottomrule
\end{tabular}
\end{wraptable}
\par\medskip

\paragraph{Even failed trajectories can provide constructive hints.}
While AutoGuide improves performance over ReAct, its gains are larger for weaker base models and often limited to relatively simple hints due to its reliance on contrastive pairs. In contrast, \MethodName{} outperforms AutoGuide by generating hints from \emph{all} available trajectories—successful or failed—rather than only paired traces. This flexibility allows \MethodName{} to extract actionable guidance even from failure-only data, leading to higher task performance. To emphasize this, we report performance on tasks where the baseline ReAct agent failed entirely, shown as darker bars in \cref{fig:main_results}.


\paragraph{Entire trajectories are not always necessary for high quality hints.}

\MethodName{} improves over its non-zooming variant by selectively surfacing the most critical steps from each trajectory. To better isolate the role of this mechanism, we additionally compare \MethodName{} to a variant that provides the hinter with the \emph{entire} execution trace—every observation, action, and think token—without any step selection, up to the maximum context length. Using GPT-5 for both the base agent and the hinter, supplying a full single trace yields only marginal gains over ReAct. In contrast, zooming over two trajectories produces substantially larger improvements on both MiniWoB++ and WorkArena-L1 (Table~\ref{tab:zoom_ablation}). These findings indicate that the hinter benefits not from receiving more context, but from receiving more \emph{informative} context: the ability to compare trajectories and focus on high-salience AXTree snapshots is essential for generating actionable hints. Since zooming is performed entirely offline, these gains come at no additional inference-time cost.

\subsection{How effective is \MethodName{} compared to documentation and human hints?}

\begin{wraptable}{r}{0.45\columnwidth}
    \centering
    \small
    \caption{Comparison of \MethodName{} against alternative hinting strategies. 
    Results are reported as average reward with standard error of $0.01$ on WorkArena-L1 
    and $0.03$ on WebArena-Lite.}
    \label{tab:alternative_hints_results}
    \begin{tabular}{lcc}
        \toprule
        \textbf{Method} & \textbf{WorkArena-L1} & \textbf{WebArena-Lite} \\
        \midrule
        \multicolumn{3}{c}{\textbf{\textsc{\scriptsize GPT-5-nano}}} \\
        ReAct & 0.41 & 0.23 \\
        Human hints & 0.43 & -- \\
        Documentation & 0.44 & 0.20 \\
        \MethodName{} & \textbf{0.48} & \textbf{0.27} \\
        \midrule
        \multicolumn{3}{c}{\textbf{\textsc{\scriptsize GPT-5-mini}}} \\
        ReAct & 0.61 & 0.32 \\
        Human hints & 0.66 & -- \\
        Documentation & 0.64 & 0.33 \\
        \MethodName{} & \textbf{0.68} & \textbf{0.34} \\
        \bottomrule
    \end{tabular}
\end{wraptable}

To assess the value of trajectory-based hints, we compare \MethodName{} against two alternative sources: platform documentation and human-authored instructions. Unlike \MethodName{}, these hints are not distilled from trajectories but taken directly from raw resources—documentation webpages or short annotator notes—and retrieved at inference time. This comparison tests whether explicit guidance can match or exceed the utility of trajectory-derived hints.

For documentation, we collected platform-specific materials: ServiceNow for WorkArena-L1, and GitLab and Shopping sites for WebArena. Pages were retrieved with BM25 using the task goal as the query, and the top-ranked passages were provided directly to the agent as hints (see \cref{sec:doc_search} for details). Human hints were prepared only for WorkArena-L1: we curated concise notes for 16 particularly challenging goals, covering all task types while focusing on cases where automated hinting failed. In both baselines, the retrieved content was used as a direct substitute for trajectory-based hints, not in combination. Results of these comparisons are reported in \cref{tab:alternative_hints_results}, with details of the human hint collection in \cref{sec:human_hints}.
External resources can substitute for trajectory-based hints, but with notable trade-offs. Documentation retrieval scales easily and provides modest gains, though its utility depends heavily on manual quality and often yields only partially relevant context. Human hints (limited to 16 curated goals), while effective are expensive to obtain and hard to scale. Overall, both baselines help bridge knowledge gaps, but \MethodName{} is more practical: it automatically produces reusable hints from offline traces without relying on manuals or human annotation.

\subsection{Can \MethodName{} generalize out-of-task?} 

\begin{figure}[ht]
    \centering
    \begin{subfigure}{0.48\textwidth}
        \centering
        \includegraphics[width=\linewidth]{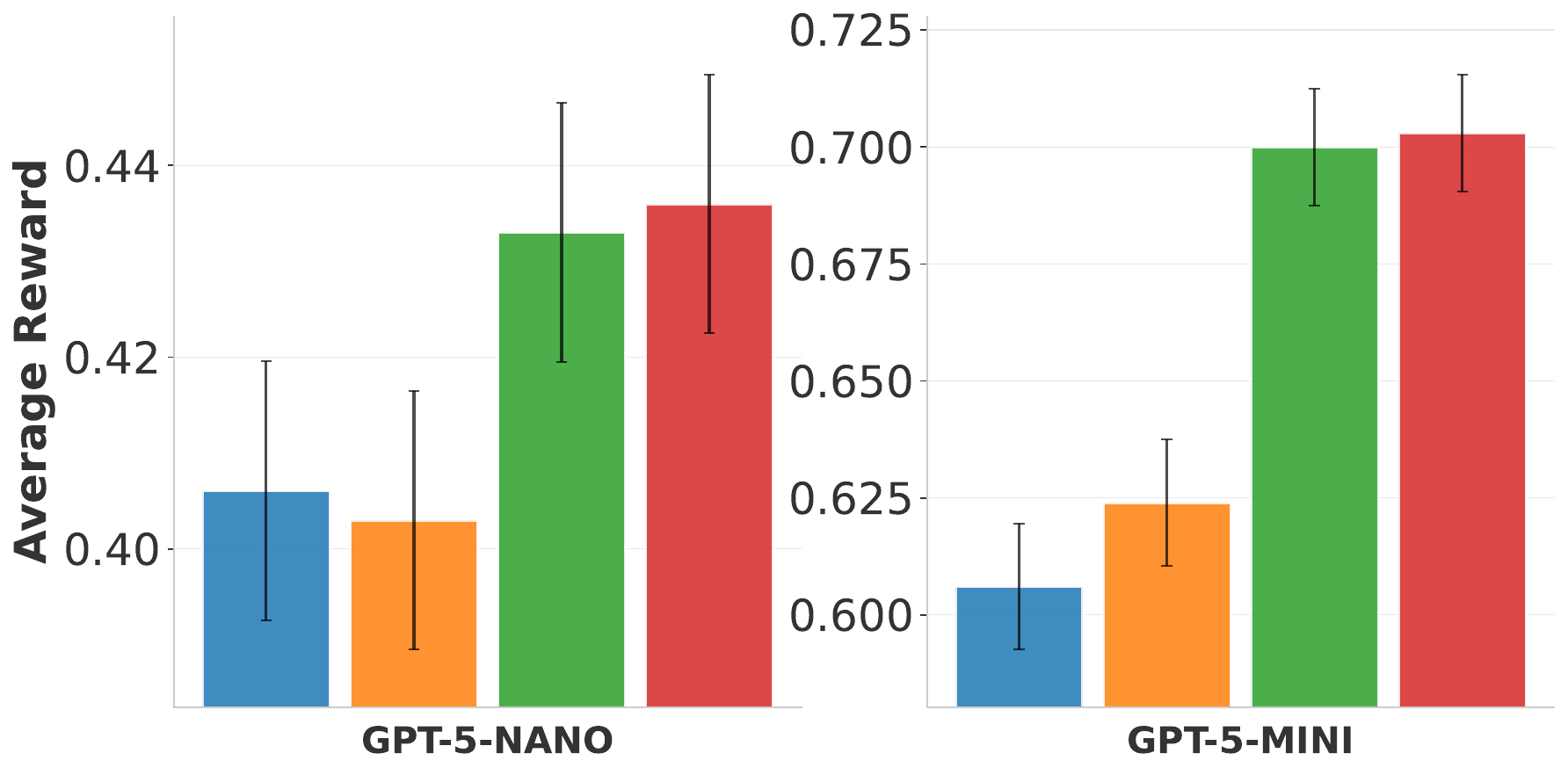}
        \caption{WorkArena-L1}
        \label{fig:ood_results_a}
    \end{subfigure}
    \hfill
    \begin{subfigure}{0.48\textwidth}
        \centering
        \includegraphics[width=\linewidth]{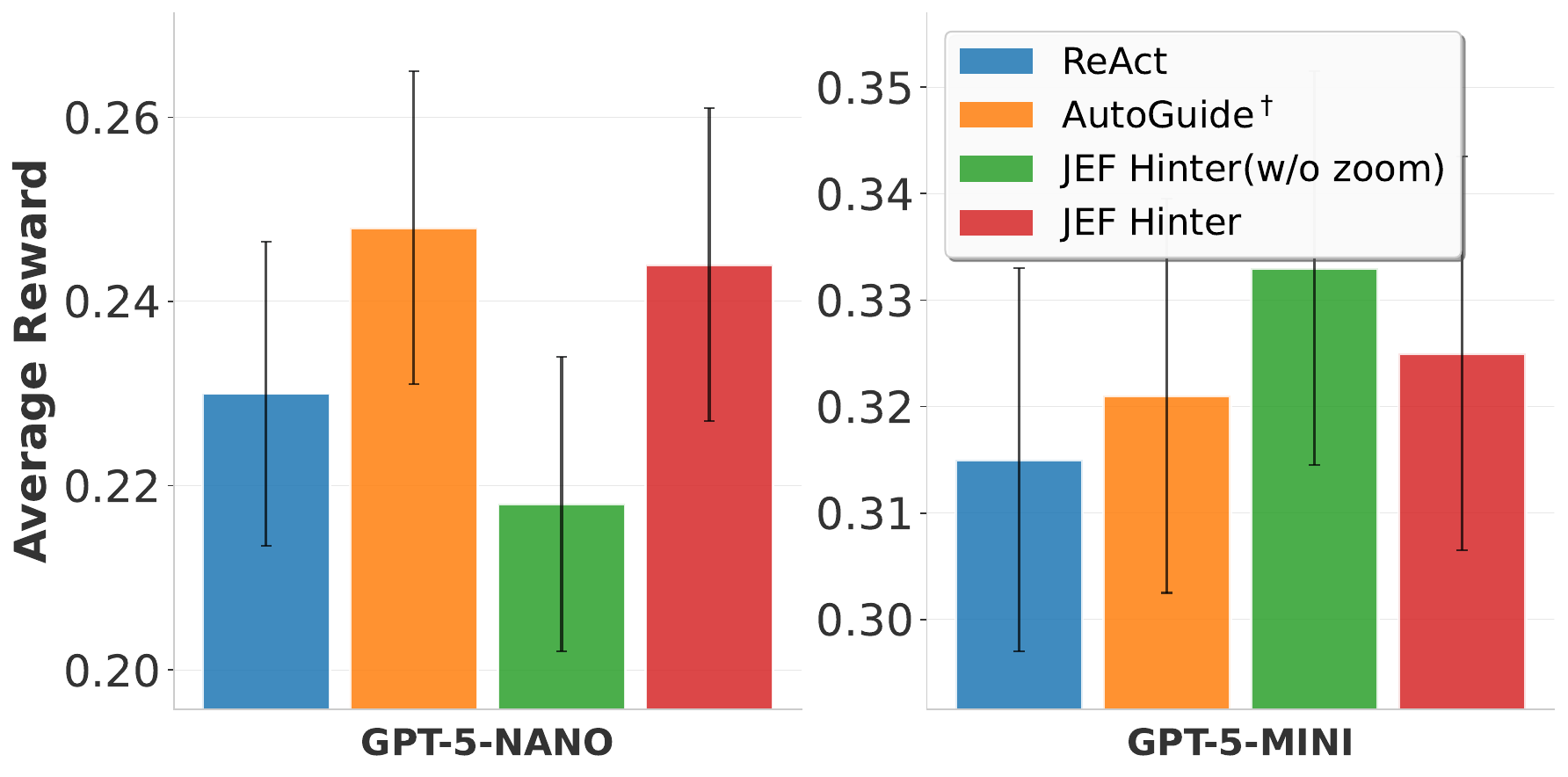}
        \caption{WebArena-Lite}
        \label{fig:ood_results_b}
    \end{subfigure}
    \caption{Out-of-task generalization performance on WorkArena-L1 and WebArena-Lite using two base models with \textbf{GPT-5-mini} as the base for the hinter model.}
    \label{fig:ood_results}
\vspace{-2mm}
\end{figure}

To assess out-of-task generalization, we remove the source task used to generate hints from the retrieval pool. The retriever must then select the most relevant hints by matching the current task goal against the remaining database entries. As shown in \cref{fig:ood_results}, \MethodName{} sustains competitive performance under this setting, indicating that trajectory-derived hints can transfer beyond the tasks they were trained on. On WorkArena-L1, we still observe clear gains over both ReAct and AutoGuide, while on WebArena-Lite, all methods perform within the margin of noise, suggesting that this benchmark remains challenging for cross-task transfer.

Our out-of-task results demonstrate that hints can transfer within an environment when tasks share similar interaction patterns (e.g., filtering and sorting both involve list manipulation). Cross-benchmark generalization (e.g., MiniWoB++ to WorkArena) is not evaluated because hints inherently encode environment-specific structures: ServiceNow's navigation hierarchy differs fundamentally from simple MiniWoB++ interfaces. This specificity is not a limitation but a design choice—just as human expertise and documentation are environment-specific. The value of \MethodName{} lies in \textit{accumulating reusable knowledge within recurring workflows}, where the same or similar tasks appear repeatedly, and systematic learning from failures prevents error repetition. For organizations deploying agents in specific platforms (e.g., enterprise software, e-commerce sites), this within-environment adaptation is the primary use case.

\subsection{Analysis \& Discussion}

\textbf{How does the size of the hinter model affect performance?}  

\begin{figure}[ht]
    \centering
    \begin{subfigure}{0.28\textwidth}
        \centering
        \includegraphics[width=\linewidth]{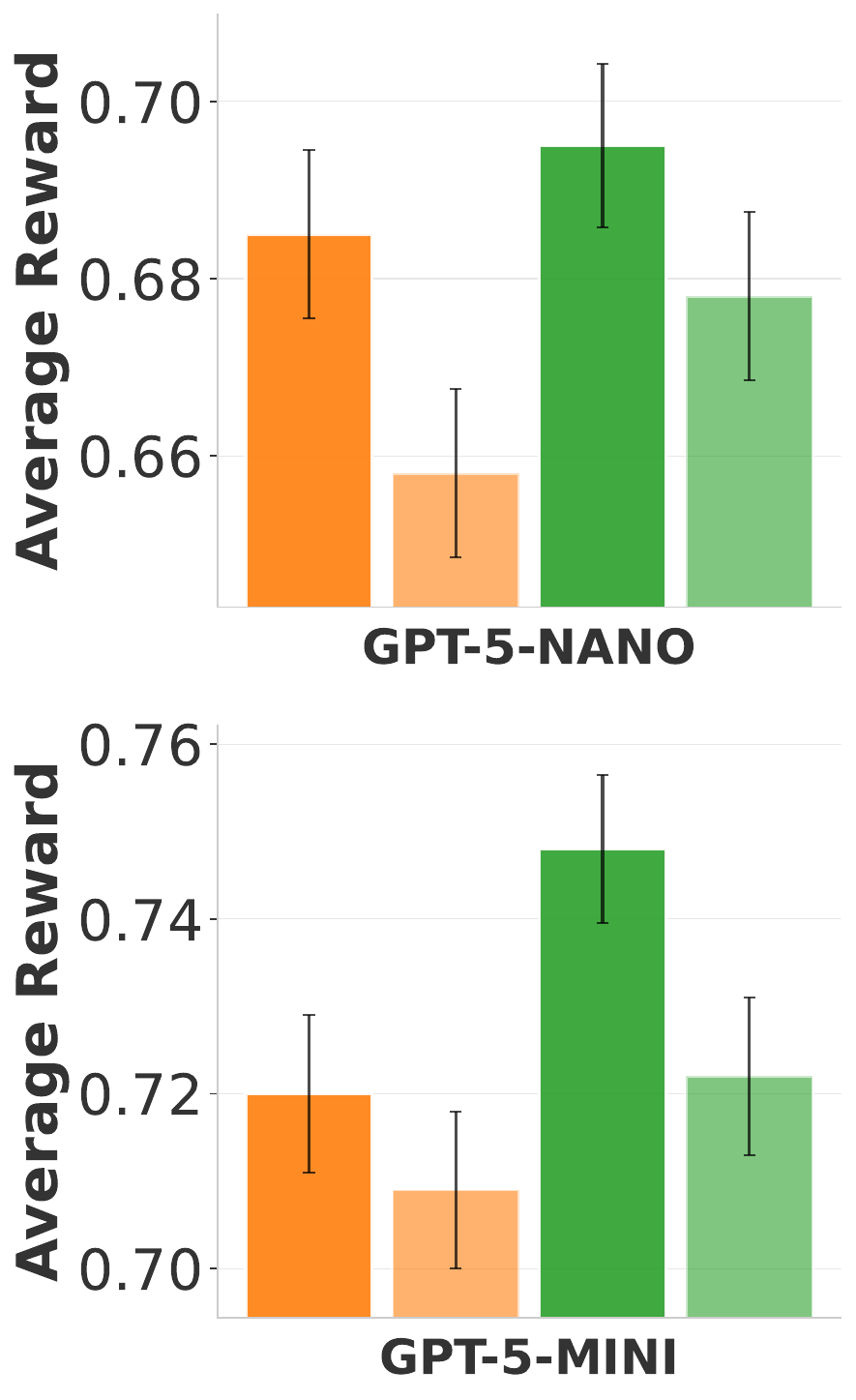}
        \caption{Miniwob++}
    \end{subfigure}
    \hfill
    \begin{subfigure}{0.28\textwidth}
        \centering
        \includegraphics[width=\linewidth]{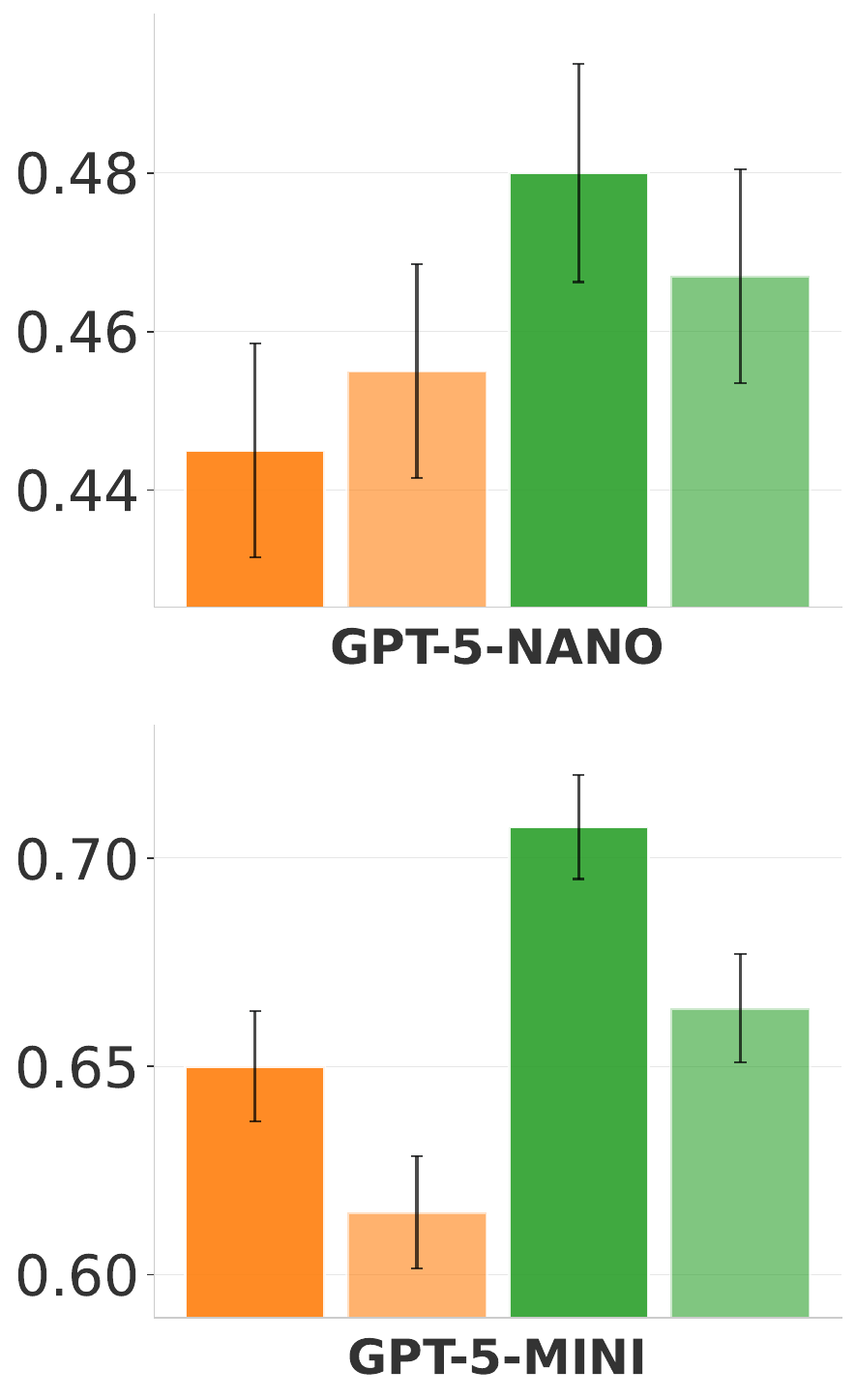}
        \caption{WorkArena-L1}
    \end{subfigure}
    \hfill
    \begin{subfigure}{0.28\textwidth}
        \centering
        \includegraphics[width=\linewidth]{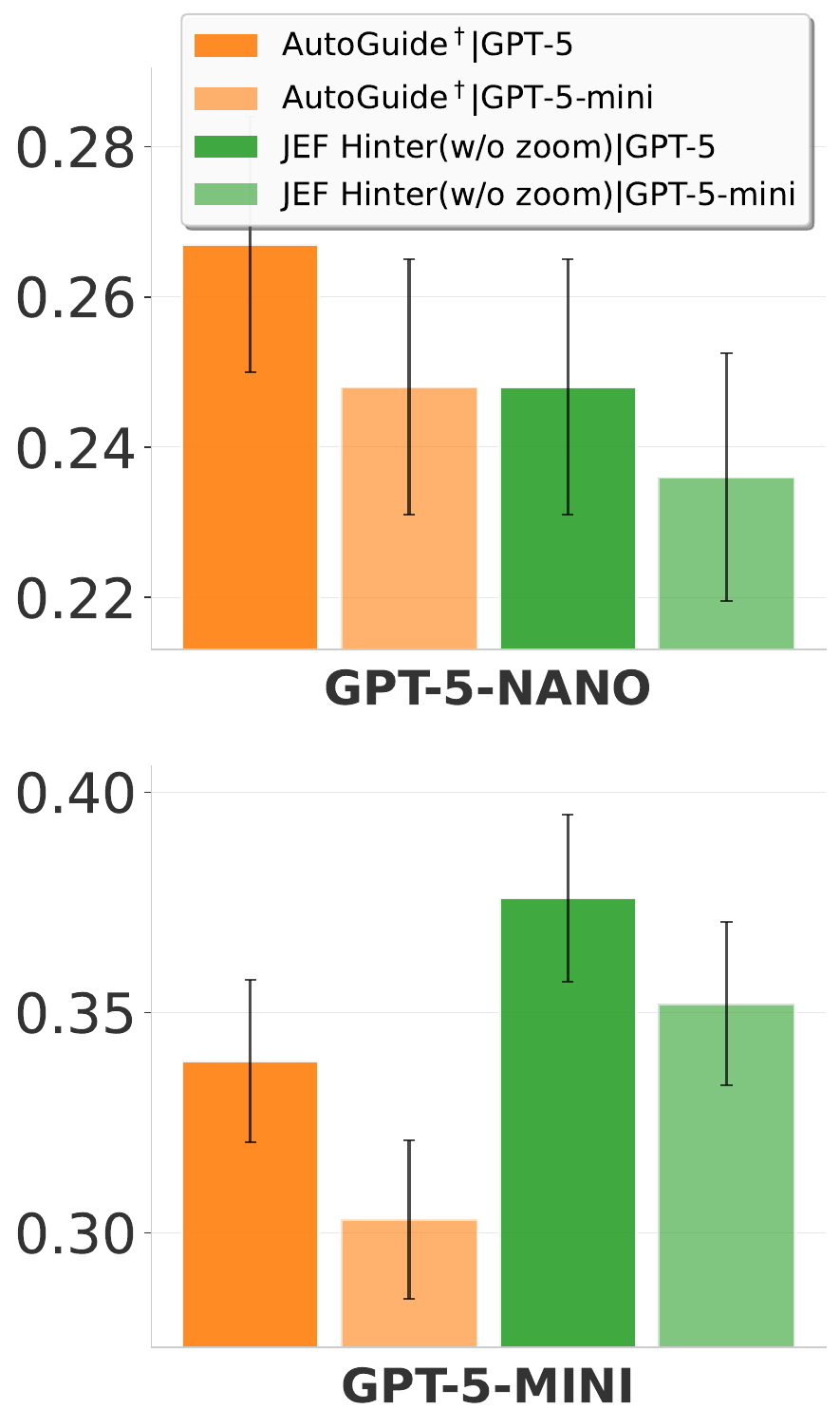}
        \caption{WebArena-Lite}
    \end{subfigure}
    \caption{Comparison of hinter models (GPT-5-mini vs.\ GPT-5) on MiniWoB++, WorkArena-L1, and WebArena-Lite. Larger hinters generally provide higher-quality hints, with the biggest gains on complex, long-horizon tasks.}

    \label{fig:hinter_models}
\end{figure}

\Cref{fig:main_results} showed that GPT-5-mini can already serve as a capable hinter for both GPT-5-nano and itself. To isolate the effect of capacity, we ablate the hinter model from GPT-5-mini to GPT-5. As shown in Figure~\ref{fig:hinter_models}, the larger hinter generally produces higher-quality hints, translating into stronger downstream performance. Gains are most pronounced on complex, long-horizon tasks such as WorkArena-L1($+5\%$), where fine-grained context understanding and precise hint phrasing matter most. On simpler tasks like MiniWoB++ ($+2\%$), the advantage narrows, suggesting that larger hinters are particularly useful when reasoning demands are high. Thus, scaling the hinter model improves performance but introduces a clear trade-off between quality and computational cost.

\paragraph{Qualitative analysis.}
Case studies illustrate how \MethodName{}’s hints intervene precisely at the decision points that previously caused failures, directly correcting the agent’s reasoning and enabling successful task completion.

\begin{tcolorbox}[
  title=Qualitative Examples of Retrieved Hints,
  colback=gray!5,
  colframe=gray!60,
  boxrule=0.4pt,
  arc=2pt,
  left=6pt,right=6pt,top=4pt,bottom=4pt,
  fonttitle=\bfseries,
  before skip=6pt, after skip=6pt
]

\textbf{MiniWoB++.}
In \texttt{click-scroll-list}, the agent must select multiple items (e.g., Bermuda and Saint Lucia) and click Submit.
Without hints, the agent often clicks items sequentially without holding a modifier key, causing earlier selections to be cleared.
\MethodName{} retrieves a hint explicitly recommending multi-selection via holding \texttt{Ctrl} (or \texttt{Cmd} on Mac) while clicking each required item, which resolves this recurring failure mode.
See Appendix~\ref{appendix:hint_examples} for full trace.

\medskip
\textbf{WorkArena-L1.}
In \texttt{filter-navigation} tasks, ReAct frequently fails by using the wrong search context (e.g., global search or \emph{Workspaces} filters) and/or by clicking before the Application Navigator menu has expanded, leading to repeated loops.
\MethodName{} provides an explicit hint to use the Navigator’s \emph{All} menu, type the application name in the correct filter box, wait for the menu to populate, and then click the target module.
This guidance helps the agent reliably reach the intended \emph{Active} module while avoiding wasted actions.

\medskip
\textbf{WebArena-Lite.}
In Shopping Admin tasks that require aggregating cancellations over the full order history, the baseline often fails by inspecting only the first page and overlooking default filters (e.g., date or keyword constraints), which undercounts results.
\MethodName{} follows a structured hint that applies a \emph{Canceled} status filter, clears irrelevant constraints, sorts/group by \emph{Bill-to Name}, and uses pagination to identify the largest group.
This enables correct end-to-end completion; Figure~\ref{fig:hint_leveraged_by_agent_webarena} visualizes how the agent uses the hint.

\end{tcolorbox}

\paragraph{Case Study for Zooming Mechanism}(WorkArena-L1 \texttt{sort-hardware-list} task). This task demonstrates the critical impact of zooming: \MethodName{} reaches only 10\% success without zooming but 70\% with zooming enabled, using the same base agent. The gain comes entirely from higher-quality hints. Zooming surfaced four decisive steps from both successful (Steps 4, 8) and failed (Steps 14, 16) trajectories, capturing the minimal causal backbone: "open filter panel → add sort rows → configure fields/directions → apply sort." This enables the hinter to produce crisp, structured hints like "expand the filter panel, click 'Add Sort,' choose fields in priority order, select directions, then click 'Run filter' to apply." In contrast, non-zoomed traces generate unfocused hints like "open Personalize List, add fields, wait for headers to load, then click headers or use Actions $>$ Sort," which fail to capture the canonical workflow. This improvement demonstrates why zooming is essential for complex, long-horizon web tasks.

\section{Conclusion}
We present \MethodName{}, an agentic system that distills large offline traces into short, retrievable hints that help agents overcome common failure modes. \MethodName{} uses a zooming module to identify critical decision points in long trajectories. A reflection step then distills these segments into reusable strategies and pitfalls. The resulting hints are compact, transparent, and easily injected at inference without fine-tuning. Experiments on MiniWoB++, WorkArena-L1, and WebArena-Lite show improvements over strong baselines, including gains in both out-of-goal and out-of-task generalization. Ablations further highlight how retrieval design, hinter capacity, and the inclusion of failed trajectories shape downstream performance, offering actionable insights for future applications. We view this work as a step toward data-centric adaptation of LLM agents, where past trajectories, documents, and human instructions are systematically mined into reusable knowledge for more robust and resilient decision-making.


\paragraph{Reproducibility Statement.} 
The reproducibility of experiments on web agents poses several challenges, as it relies on a software stack for hosting the environment server and the backend of the web agent. To address this, we rely on AgentLab and BrowserGym\citep{dechezelles2025browsergymecosystemwebagent}, a framework designed for evaluating agents with reproducibility in mind. Among other features, the version of all installed packages used during the experiments is saved in the experiment results. In addition to open-sourcing our code, we will also provide all experiment traces as provided by AgentLab. In the meantime, an anonymized codebase is provided in the supplementary materials.

For the reproducibility of our method,
Section~\ref{sec:method}, which provides a detailed description of the \MethodName{} framework, while Section~\ref{sec:exp_setup} specifies benchmarks, baselines, and evaluation protocols. Appendix~\ref{appendix:system_prompts} includes the full prompts used for hint generation and retrieval, Appendix~\ref{sec:doc_search} describe documentation and human hint collection procedures, and Appendix~\ref{appendix:hint_examples} provides case studies with reasoning traces. All datasets (MiniWoB++, WorkArena-L1, and WebArena-Lite) are publicly available, and we include details of our offline data collection and augmentation pipeline in Section~\ref{sec:exp_setup}.

\bibliography{servicenow}
\bibliographystyle{servicenow}

\newpage
\appendix
\section{Documentation Search as Hints for LLM Agents}\label{sec:doc_search}

We explore the use of documentation search as a hinting mechanism, enabling agents to retrieve relevant knowledge directly from official platform resources. Specifically, we scrape documentation from ServiceNow\footnote{https://www.servicenow.com/docs/} for WorkArena-L1, and from GitLab\footnote{https://docs.gitlab.com/} and shopping websites\footnote{https://experienceleague.adobe.com/en/docs/commerce-admin/user-guides/home} for WebArena. Each webpage is converted into a cleaned markdown file with a structured header that records metadata such as the page title, summary, keywords, and breadcrumbs.

\paragraph{Experimental Setup} To evaluate how best to retrieve relevant hints, we explore three complementary design dimensions:
\begin{itemize}
    \item \textbf{Retrieval method}. We compare sparse retrieval with BM25 \citep{robertson1995okapi} against dense retrieval using pretrained embeddings \citep{karpukhin2020dense}.
    \item \textbf{Query formulation}. We test using the raw task goal as the query versus prompting the LLM to generate a more specific query from the current task context. This comparison mirrors episode-level hints versus step-level hints.
    \item \textbf{Granularity of retrieval}. We contrast retrieving full documentation pages with retrieving structured chunks. In the chunked setting, we align snippets with the markdown hierarchy, treating each section as an independent unit without overlap.
\end{itemize}

Information about the extracted documentation webpages can be found in Table \ref{tab:documentation_stats}.

We evaluate configurations on WorkArena-L1 using GPT-5-mini as the base model. Sparse retrieval is implemented with \texttt{bm25s} \cite{bm25s}, while dense retrieval uses \texttt{embeddinggemma-300m} \cite{vera2025embeddinggemma} with Faiss \cite{douze2024faiss}. For reformulated queries, GPT-5-mini generates context-aware search strings. To ensure fairness across setups, we fix the retrieval depth: the full-page setting returns the top 3 pages, and the chunked setting returns the top 5 section-level snippets.

\begin{table}[ht]
    \centering
    \begin{minipage}{0.45\linewidth}
        \centering
        \caption{Documentation Corpus Statistics: Number of Pages and Chunks per Platform}
        \label{tab:documentation_stats}
        \begin{tabular}{lll}
        \toprule
            \textbf{Platform} & \textbf{\# Pages} & \textbf{\# Chunks} \\
            \midrule
            ServiceNow & 60,967 & 287,271 \\
            GitLab & 2,654 & 35,470 \\
            Shopping & 598 & 4,010 \\
        \bottomrule
        \end{tabular}
    \end{minipage}%
    \hfill
    \begin{minipage}{0.52\linewidth}
        \centering
        \caption{Comparison of Documentation Search Settings for Web-Browsing Agents}
        \label{tab:documentation_hinting_results}
        \begin{tabular}{llll}
        \toprule
            \textbf{Search} & \textbf{Query} & \textbf{Document} & \\
            \textbf{Type} & \textbf{Type} & \textbf{Type} & \textbf{Reward} \\

            \midrule
            N/A & N/A & N/A & 0.61 \\
            Sparse & Goal & Full  & \textbf{0.64} \\
            Sparse & Goal & Chunk & \underline{0.63} \\
            Sparse & LLM  & Full  & 0.62 \\
            Sparse & LLM  & Chunk & \underline{0.63} \\
            Dense  & Goal & Full  & 0.60 \\
            Dense  & Goal & Chunk & 0.59 \\
            Dense  & LLM  & Full  & 0.62 \\
            Dense  & LLM  & Chunk & 0.61 \\
        \bottomrule
        \end{tabular}
    \end{minipage}
\end{table}

\paragraph{Results} The ablation results across these configurations are reported in Table~\ref{tab:documentation_hinting_results}. Overall, we find that \textbf{a simple retrieval framework is highly competitive}. Using BM25 with the task goal as the query and retrieving full pages achieves performance on par with more complex dense retrieval and LLM query reformulation setups. This configuration is also faster and easier to implement, making it a strong baseline for documentation-based hinting. While advanced retrieval pipelines provide only marginal gains, simplicity and efficiency often suffice for supplying LLM agents with actionable documentation hints. Dense retrieval in particular underperforms, likely due to embeddings being less attuned to domain-specific technical terminology.

\paragraph{Discussion} In most cases, we find that documentation pages are not a reliable source of instructions for navigating complex user interfaces. Unlike tutorials designed for end-users, documentation rarely specifies how to perform low-level interactions such as clicking, scrolling, or filling forms. As a result, retrieved passages often contain information that is only tangentially related to the task at hand. Encouragingly, the agent is generally able to disregard irrelevant context and maintain a similar level of performance, even if individual successes and failures shift across tasks. In other words, documentation hints can occasionally distract the agent, but the net effect on performance is largely stable when the provided context is unhelpful.

The impersonation task stands out as the most notable case where documentation significantly improves performance. Without hints, GPT-5-mini frequently refuses to act, interpreting "impersonation" as unsafe rather than recognizing it as a legitimate ServiceNow feature. This reflects an alignment artifact, where the model overgeneralizes safety constraints to benign enterprise contexts. Providing the impersonation documentation resolves this issue, enabling successful execution. This example highlights the dual benefit of documentation retrieval: it can both supply missing procedural knowledge and clarify task intent in ways that help override misaligned safety refusals. In contrast, tasks such as filtering and sorting show degradation primarily due to skill-based errors, underscoring that documentation hints are most impactful in cases where alignment conflicts, rather than procedural gaps, are the limiting factor.

\paragraph{Limitations} A key limitation of documentation-based hinting is its reliance on the availability of high-quality resources. Within WebArena, only GitLab and Shopping/Shopping Admin tasks are supported by relevant documentation, and even these are far less comprehensive than ServiceNow’s materials in WorkArena-L1. Other platforms, such as OpenStreetMap and Postmill, offer little to no user-facing documentation. As also noted by \cite{song2024beyond}, the breadth and quality of documentation directly affect agent performance, particularly for tasks requiring API-level interaction. This underscores that documentation-based approaches may not generalize uniformly across platforms.

\section{Human Hint Collection}
\label{sec:human_hints}

\begin{figure}[!ht]
    \centering
    \includegraphics[width=\textwidth]{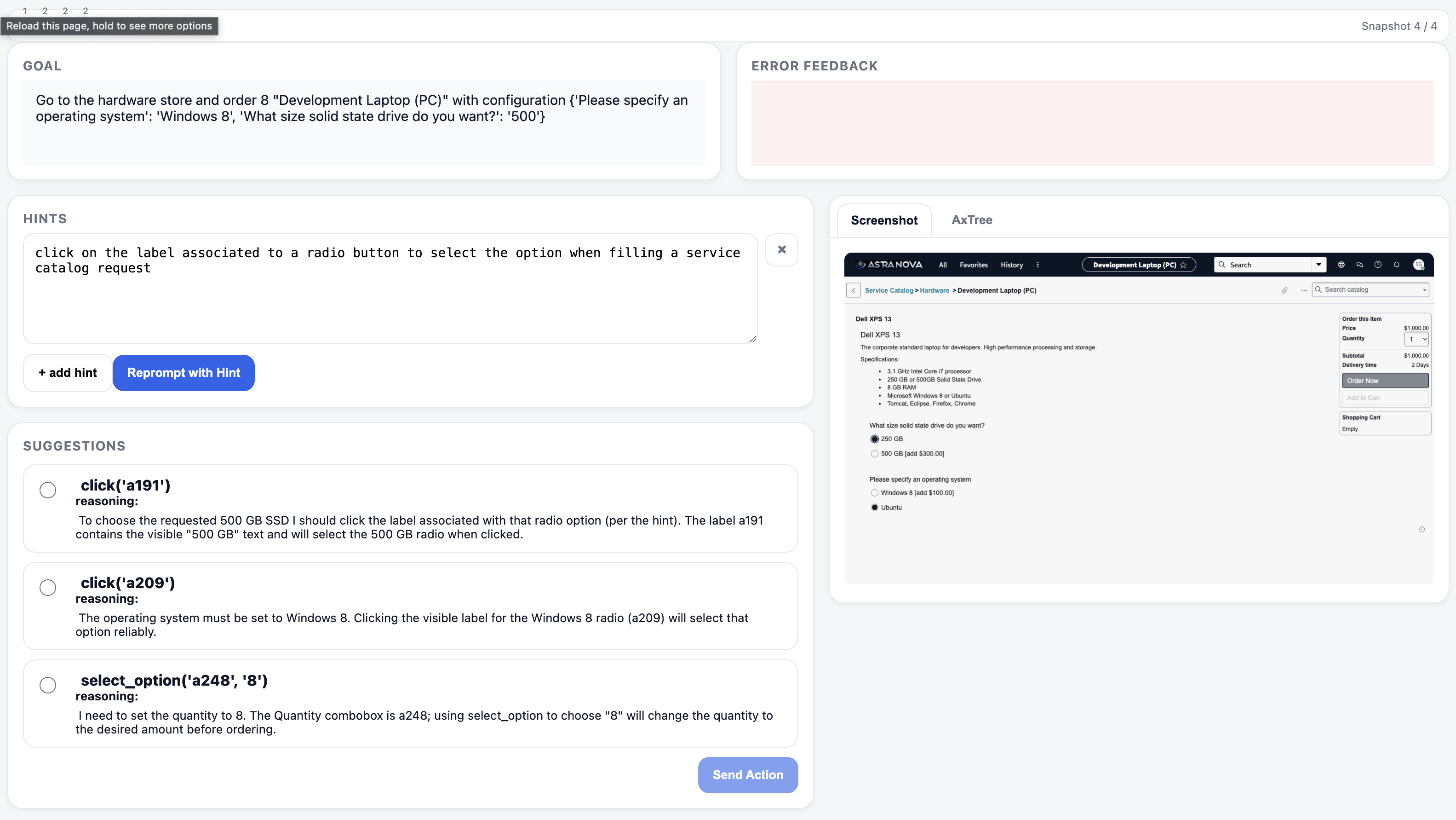}
    \caption{Interactive labeling interface used for human hint collection. Annotators selected actions from a model–generated list, or provided free–form hints when the desired action was missing. The updated candidates were then re–evaluated until the task was completed.}
    \label{fig:hint_labeling_ui}
\end{figure}

To gather high–quality hints from humans, we designed an interactive annotation interface that places the human annotator in the loop of action selection. At each step of a task, the model proposes a list of candidate actions. If the correct action is among them, the annotator simply selects it. Otherwise, the annotator can provide a free–form hint that guides the model toward the desired action. The model then regenerates a new set of candidate actions conditioned on this hint, and the cycle continues until the task is successfully completed. This iterative process ensures that we collect both the final action sequence and, importantly, the intermediate natural language hints produced by humans. \cref{fig:hint_labeling_ui} presents the labeling UI used to collect human hints.

The hints serve to make explicit the reasoning behind otherwise opaque choices. For example, when filtering a table, annotators often wrote instructions such as: \textit{click on the gridcell that says "-- choose field --" to pick Category} or \textit{let's do one condition at a time. Click on "choose field" so that we can select Assigned to}. Similarly, when filling a multi–tab form, annotators specified \textit{to set the assignment group, click on the look up icon}. These hints capture localized decision strategies and offer the model additional guidance beyond raw demonstrations. By collecting such hints alongside trajectories, we create a resource that directly encodes human teaching signals and can be reused to improve model alignment with task–specific interaction patterns.


\section{System prompts}\label{appendix:system_prompts}
\subsection{Step selection }\label{appendix:zooming_prompt}

\begin{tcolorbox}[colback=white,colframe=black,title={Prompt for Step Selection}]
You are a trace summarizer. Given the following execution trace, identify the step or steps that are most important for understanding success or failure. Return the step numbers (starting from 1) and a brief reason why they are important.\par

=== EXECUTION TRACE ===\par
Goal: \texttt{<TASK GOAL>}\par
Step 1: ...\par
Step 2: ...\par

=== STEP SELECTION CRITERIA ===\par
Look for steps that are critical because they:\par
1. Represent a key decision point or branching moment\par
2. Show a common mistake that could be avoided\par
3. Demonstrate a successful strategy or pattern\par
4. Involve important UI elements or context clues\par
5. Show timing or sequence dependencies\par
6. Represent the moment where success/failure was determined\par

=== STEP SELECTION ===\par
List the most important step numbers (comma separated) and a brief reason for each.\par
IMPORTANT: Do not repeat the same step number. Select 1--2 critical steps that provide the most valuable insights for generating actionable hints.\par

=== THINKING PROCESS ===\par
Before selecting the most important steps, think through:\par
1. Which steps represent critical decision points?\par
2. Which steps show avoidable mistakes?\par
3. Which steps demonstrate successful strategies?\par
4. Which steps involve important UI/context clues?\par
5. Which steps show timing or sequence dependencies?\par
6. Which steps mark where success/failure was determined?\par

Think step by step and analyze carefully before making your selection.
\end{tcolorbox}

\subsection{Step-Sequence Hinting}\label{appendix:step_sequence_prompt}
\begin{tcolorbox}[colback=white,colframe=black,title={Prompt for Step-Sequence Hint Generation}]
Task: \texttt{<TASK NAME>}\par
\textbf{=== STEP SEQUENCE ANALYSIS (}<N\textbf{ consecutive steps) ===}\par
Goal: \texttt{<TASK GOAL>}\par
Step i: \texttt{Observation(s), Agent's reasoning, Action taken, Error encountered, Current reward}\par
Step i+1: \texttt{Observation(s), Agent's reasoning, Action taken, Error encountered, Current reward}\par
\ldots\par

=== STEP-SEQUENCE HINT GENERATION ===\par
Based on the sequence of \texttt{<N>} consecutive steps above, provide a concise, actionable hint that explains:\par

1. What the agent accomplished across these steps.\par

2. What the agent should do next based on the context.\par

3. How to recognize when this sequence is needed.\par

4. Common mistakes to avoid during this sequence.\par

=== STEP-SEQUENCE GUIDANCE ===\par
Focus on:\par
-- What changed in the environment across these steps.\par
-- What the agent learned or accomplished.\par
-- The next logical action.\par
-- How to recognize the right moment for that action.\par
-- The pattern or workflow this sequence represents.\par

\textbf{Include the full Hint Requirements'' and Output Format'' as in \Cref{appendix:hint_generation_prompt}.}\par
\end{tcolorbox}
\subsection{Hint Generation}\label{appendix:hint_generation_prompt}
\begin{tcolorbox}[breakable, colback=white, colframe=black,title={Prompt for Hint Generation (Single / Multi-Trace)}]
\textbf{System role}\par
You are a hint generation expert. You MUST respond using the structured format with \texttt{<think>}, \texttt{<topic>}, and \texttt{<hint>} tags. Use the \texttt{<think>} section for thorough analysis (200--800 words) and the \texttt{<hint>} section for concise, actionable guidance (under 256 tokens, single line).\par

=== INPUT ===\par
Task: \texttt{<TASK NAME>}\par
Goal: \texttt{<TASK GOAL>}\par
(Optional) Documents/Instructions: \texttt{<SHORT SNIPPETS OR NONE>}\par
\textbf{Execution trace(s):}\par
Step 1: \texttt{Observation(s), Agent's reasoning, Action taken, Error encountered, Current reward}\par
Step 2: \texttt{Observation(s), Agent's reasoning, Action taken, Error encountered, Current reward}\par
\ldots\par
(Repeat for each provided trace when multiple traces are given)\par

=== HINT REQUIREMENTS ===\par
IMPORTANT: Keep your hint SHORT and write it as a SINGLE LINE without line breaks.\par
Focus on:\par
-- Common pitfalls or errors to avoid\par
-- Specific strategies that work well\par
-- Important details and UI cues to pay attention to\par
-- Step-by-step guidance if multiple actions are required\par

=== ENHANCED REQUIREMENTS ===\par
\textbf{Generalizability}\par
-- Make hints general enough to apply to similar tasks, not just this specific instance.\par
-- DO NOT include: specific usernames, literal task content strings, element IDs like [123], domain-specific secrets.\par
-- DO include: reusable UI patterns (buttons, links, form fields), common workflows, robust strategies.\par

\textbf{Specificity \& Actionability}\par
-- Use exact UI text only when it represents common patterns (e.g., button labels like 'Submit').\par
-- Specify element types and positions when relevant (e.g., button at the bottom of the form).\par
-- Provide clear step ordering when multiple actions are needed.\par

\textbf{Structure \& Length}\par
-- Hint under 256 tokens, single line, no line breaks.\par
-- Focus on what to do, not why it works.\par
-- Use single quotes (') and \emph{never} double quotes (") in the hint.\par

\textbf{Topic Tag}\par
-- Always provide one short sentence describing the applicability topic inside \texttt{<topic>} tags (e.g., \texttt{filtering the table}, \texttt{multi-tab form filling}).\par
-- If a line \texttt{SUMMARIZATION: <summarization>} is present in the input, incorporate it into the \texttt{<topic>} description.\par

Known applicability topics: \texttt{<TOPIC LIST IF AVAILABLE>}\par
=== OUTPUT FORMAT ===\par
\texttt{<think>}\par
Your reasoning about the traces, patterns, decisive steps, and reusable strategies (200--800 words).\par
\texttt{</think>}\par

\texttt{<topic>}\par
One short sentence describing the general task topic (e.g., \texttt{filtering the table}).\par
\texttt{</topic>}\par

\texttt{<hint>}\par
A single-line, concise, actionable hint under 256 tokens (use single quotes, no line breaks).\par
\texttt{</hint>}\par

=== THINKING SECTION GUIDANCE ===\par
-- Analyze the execution traces in detail.\par
-- Identify key patterns, mistakes, and successful strategies.\par
-- Explain (in \texttt{<think>}) why certain approaches work or fail.\par
-- Consider multiple perspectives and edge cases.\par
-- Aim for 200--800 words of thoughtful analysis.\par

=== HINT SECTION GUIDANCE ===\par
-- Focus on the most critical action(s) the agent should take next.\par
-- Avoid lengthy explanations or multiple examples.\par
-- Prioritize what to do; keep it executable.\par
-- Keep it under 256 tokens; use single quotes only.\par

=== EXAMPLES (GOOD) ===\par
\textbf{Example 1 - Navigation:}\par
\texttt{<think>}\par
Looking at the execution traces, the agent often fails by using the global search instead of the left-side Application Navigator. Successful runs type into 'Filter/Filter navigator' and click module links after the app expands. Repeated clicks on admin menus are unnecessary; the key is filtering in the left panel and then selecting the specific module entry once visible.\par
\texttt{</think>}\par
\texttt{<hint>}\par
Use the Application Navigator (left panel) with the 'Filter/Filter navigator' input to find and open modules; do not use the global search bar at the top.\par
\texttt{</hint>}\par

\textbf{Example 2 - Form Submission}\par
\texttt{<think>}\par
Agents fail when expecting a 'Submit' label; successful runs click whichever action completes the flow ('Save', 'Create', or 'Submit'). Enter does not submit; explicit clicks are required.\par
\texttt{</think>}\par
\texttt{<hint>}\par
At the bottom of the form, click the action button that completes the flow (e.g., 'Save', 'Create', or 'Submit') instead of pressing Enter.\par
\texttt{</hint>}\par

=== EXAMPLES (BAD) ===\par
-- Click the button with ID [123] to submit the form. (too specific)\par
-- Enter 'john.doe@email.com
' in the email field. (too specific)\par
-- This task requires careful attention to detail. (too vague)\par
-- The agent should understand the context before proceeding. (explanatory, not actionable)\par
-- Click the "Submit" button to continue. (uses double quotes)\par
\end{tcolorbox}
\subsection{Two-Trace Comparison (Desired vs.\ Undesired)}\label{appendix:two_trace_prompt_full}
\begin{tcolorbox}[colback=white,colframe=black,title={Prompt for Two-Trace Comparison (Desired vs.\ Undesired)}]
You will be provided with a \textbf{desired (successful)} and an \textbf{undesired (failed)} trajectory for the same task. Identify the \emph{first} action where they diverge, explain why it leads to success vs.\ failure, and produce a general, reusable hint.\par

=== INPUT ===\par
Task: \texttt{<TASK NAME>}\par
Goal: \texttt{<TASK GOAL>}\par

--- \textbf{Desired trajectory} ---\par
Step 1: \texttt{Observation(s), Agent's reasoning, Action taken, Error encountered, Current reward}\par
Step 2: \texttt{Observation(s), Agent's reasoning, Action taken, Error encountered, Current reward}\par
\ldots\par

--- \textbf{Undesired trajectory} ---\par
Step 1: \texttt{Observation(s), Agent's reasoning, Action taken, Error encountered, Current reward}\par
Step 2: \texttt{Observation(s), Agent's reasoning, Action taken, Error encountered, Current reward}\par
\ldots\par

\textbf{SUMMARIZATION:} \texttt{<ONE-LINE CONTEXT SUMMARY IF AVAILABLE>}\par

=== COMPARISON GUIDANCE ===\par

1. Identify the first differing action and its local context.\par

2. Explain (in \texttt{<think>}) why one path succeeds and the other fails.\par

3. Derive a general rule that applies beyond this instance; avoid task-specific literals.\par

4. Follow the successful (desired) trajectory; do not invent steps absent from it.\par

=== OUTPUT FORMAT ===\par
\texttt{<think>}\par
Analysis of the first divergence, its effect on progress, UI/context cues to detect it, and a reusable rule (200--800 words).\par
\texttt{</think>}\par

\texttt{<topic>}\par
Short applicability topic (e.g., \texttt{using the application navigator vs.\ global search}).\par
\texttt{</topic>}\par

\texttt{<hint>}\par
Single-line, general, actionable guidance under 256 tokens; preferably in the form: When \textless status\textgreater, do \textless action\textgreater{} or Avoid \textless pitfall\textgreater{} and instead \textless action\textgreater{}. Use single quotes.
\par
\texttt{</hint>}\par
\end{tcolorbox}
\subsection{Step-Zoom Hinting}\label{appendix:step_zoom_prompt}
\begin{tcolorbox}[colback=white,colframe=black,title={Prompt for Step-Zoom Hint Generation}]
Task: \texttt{<TASK NAME>}\par
\textbf{=== ZOOMED-IN STEPS ===}\par
Goal: \texttt{<TASK GOAL>}\par
(For each step in the trace, include:) \par
Step k: \texttt{Observation(s), Agent's reasoning, Action taken, Error encountered, Current reward}\par
(For each step identified as important, additionally include the most informative structural view, e.g., AXTree or HTML.)\par

=== HINT GENERATION ===\par
Based on the most important step(s) above, provide a concise, actionable hint that would help an agent avoid common mistakes and succeed at this task.\par

=== STEP-FOCUSED GUIDANCE ===\par
1. Pay special attention to:\par

2. What makes this step decisive for success/failure.\par

3. The specific UI elements or context guiding the correct action.\par

4. Common mistakes at this decision point.\par

5. How to recognize when this step is needed.\par

6. The correct sequence or timing for this action.\par

\textbf{Include the full Hint Requirements'' and Output Format'' as in \Cref{appendix:hint_generation_prompt}.}\par
\end{tcolorbox}
\subsection{Dual-Trace Step-Zoom}\label{appendix:dual_step_zoom_prompt}

\begin{tcolorbox}[colback=white,colframe=black,title={Prompt for Dual-Trace Step-Zoom Analysis}]
Task: \texttt{<TASK NAME>}\par
\textbf{=== DUAL TRACE STEP ZOOM ANALYSIS ===}\par
For each trace (desired and undesired), provide:\par
-- Outcome summary (successful/failed) and Goal.\par
-- Steps with: \texttt{Observation(s), Agent's reasoning, Action taken, Error encountered, Current reward}.\par
-- Mark \textbf{IMPORTANT STEP} for the selected critical steps and include the relevant structural view (AXTree/HTML) for those steps.\par

=== DUAL TRACE HINT GENERATION ===\par
Based on the most important step(s) across both traces, provide a concise, actionable hint that helps avoid the observed failure.\par

=== DUAL TRACE STEP-FOCUSED GUIDANCE ===\par
Focus on:\par

1. Patterns emerging across both traces at critical steps.\par

2. Differences between correct and incorrect actions at those points.\par

3. The UI elements or context that disambiguate the right action.\par

4. Common mistakes at similar decision points.\par

5. How to recognize when these critical steps are needed.\par

6. The correct sequence/timing for actions at these points.\par

\textbf{Include the full Hint Requirements'' and Output Format'' as in \Cref{appendix:hint_generation_prompt}.}\par
\end{tcolorbox}
\subsection{Context Identification}\label{appendix:context_identification_prompt}

\begin{tcolorbox}[colback=white,colframe=black,title={Prompt for Context Identification (Pre-Retrieval)}]
You are a helpful assistant that identifies the context of a task based on trace information. You will see the prefix of a trajectory up to the first divergence between two traces. Summarize the current status to guide retrieval of relevant hints.\par

=== INPUT (TRACE PREFIX) ===\par
GOAL: \texttt{<TASK GOAL>}\par
Step 1: \texttt{Observation(s), Agent's reasoning, Action taken, Error encountered, Current reward}\par
Step 2: \texttt{Observation(s), Agent's reasoning, Action taken, Error encountered, Current reward}\par
\ldots (up to the first differing action)\par

=== INSTRUCTIONS ===\par
Before choosing an action, query memory/documentation by first generating a brief, general summary of the current status to help identify useful hints.\par
Return your answer as follows:\par
\texttt{<think>}chain of thought\texttt{</think>}\par
\texttt{<context>}one short sentence summary\texttt{</context>}\par

=== EXAMPLE ===\par
\texttt{<think>}\par
I have to sort by client and country. I could use the built-in sort on each column but I'm not sure if I can sort by both at the same time.\par
\texttt{</think>}\par
\texttt{<context>}\par
The user is preparing to apply multi-column sorting and needs guidance on adding the next criterion.\par
\texttt{</context>}\par
\end{tcolorbox}

\section{More results}
\textbf{How much faster is parallelized hint generation?}

\begin{wrapfigure}{r}{0.4\linewidth}
    \centering
    \includegraphics[width=\linewidth]{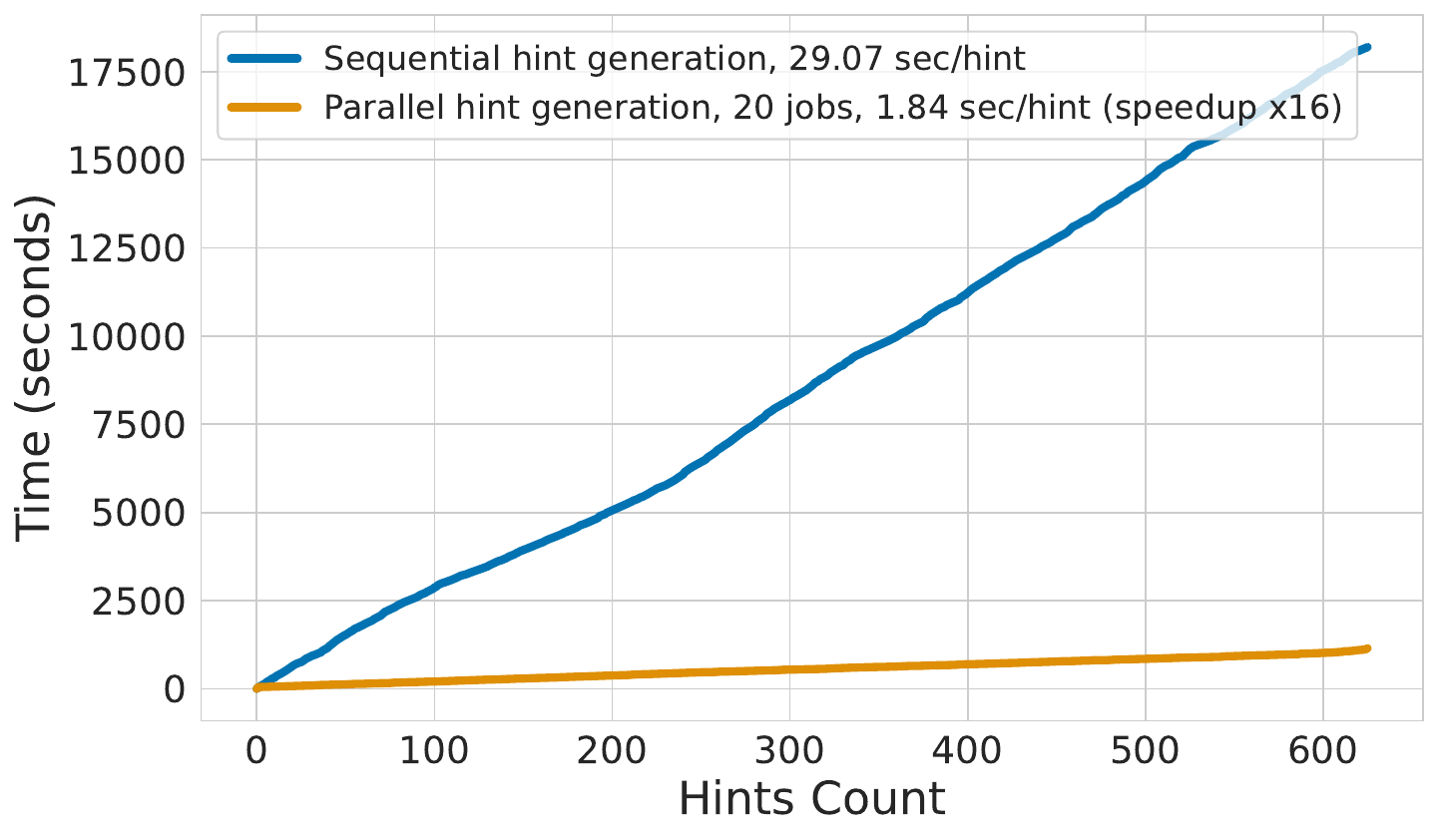}
    \caption{Parallelized hint generation.}
    \label{fig:parallel_hint_generation}
\end{wrapfigure}

The original AutoGuide guideline extraction~\cite{fu2024autoguide} module is implemented sequentially, which limits scalability. To demonstrate the efficiency of our approach, we implemented a parallelized version of hint generation that distributes trajectories across multiple workers. As shown in \cref{fig:parallel_hint_generation}, our parallel implementation achieves nearly a $20\times$ speedup over sequential hinting, enabling large-scale hint generation on complex benchmarks. This improvement makes it practical to construct diverse and comprehensive hint databases without prohibitive computational overhead. 



\newpage
\section{Hint analysis}

\subsection{Hint stats}
\begin{table}[!h]
\centering
\caption{MiniWoB++ Hint Database Statistics by Method, Base Model, and Hinter Model}
\label{tab:miniwob_hint_stats}
\resizebox{\linewidth}{!}{%
\begin{tabular}{lllccc}
\toprule
Hinter Method & Base Model & Hinter Model & Total Entries & Unique Tasks & Avg Hints/Task \\
\midrule
\multirow{4}{*}{AutoGuide-v1} 
 & gpt-5-mini & gpt-5 & 117 & 28 & 4.17 \\
 &            & gpt-5-mini-2025-08-07 & 139 & 28 & 4.96 \\
 & gpt-5-nano & gpt-5 & 157 & 36 & 4.38 \\
 &            & gpt-5-mini-2025-08-07 & 174 & 36 & 4.83 \\
\midrule
\multirow{4}{*}{\MethodName{} (w/o zoom)} 
 & gpt-5-mini & gpt-5 & 625 & 125 & 5.00 \\
 &            & gpt-5-mini-2025-08-07 & 614 & 125 & 4.91 \\
 & gpt-5-nano & gpt-5 & 625 & 125 & 5.00 \\
 &            & gpt-5-mini-2025-08-07 & 619 & 125 & 4.95 \\
\midrule
\multirow{4}{*}{\MethodName} 
 & gpt-5-mini & gpt-5 & 625 & 125 & 5.00 \\
 &            & gpt-5-mini-2025-08-07 & 618 & 125 & 4.94 \\
 & gpt-5-nano & gpt-5 & 625 & 125 & 5.00 \\
 &            & gpt-5-mini-2025-08-07 & 620 & 125 & 4.96 \\
\bottomrule
\end{tabular}%
}
\end{table}

\begin{table}[!h]
\centering
\caption{WorkArena-L1 Hint Database Statistics by Method, Base Model, and Hinter Model}
\label{tab:workarena_hint_stats}
\resizebox{\linewidth}{!}{%
\begin{tabular}{lllccc}
\toprule
Hinter Method & Base Model & Hinter Model & Total Entries & Unique Tasks & Avg Hints/Task \\
\midrule
\multirow{4}{*}{AutoGuide} 
 & gpt-5-mini & gpt-5 & 105 & 21 & 5.00 \\
 &            & gpt-5-mini-2025-08-07 & 105 & 21 & 5.00 \\
 & gpt-5-nano & gpt-5 & 155 & 31 & 5.00 \\
 &            & gpt-5-mini-2025-08-07 & 155 & 31 & 5.00 \\
\midrule
\multirow{4}{*}{\MethodName{}(w/o zoom)} 
 & gpt-5-mini & gpt-5 & 194 & 33 & 5.88 \\
 &            & gpt-5-mini-2025-08-07 & 165 & 33 & 5.00 \\
 & gpt-5-nano & gpt-5 & 188 & 33 & 5.70 \\
 &            & gpt-5-mini-2025-08-07 & 168 & 33 & 5.09 \\
\midrule
\multirow{4}{*}{\MethodName} 
 & gpt-5-mini & gpt-5 & 165 & 33 & 5.00 \\
 &            & gpt-5-mini-2025-08-07 & 164 & 33 & 4.97 \\
 & gpt-5-nano & gpt-5 & 165 & 33 & 5.00 \\
 &            & gpt-5-mini-2025-08-07 & 163 & 33 & 4.94 \\
\bottomrule
\end{tabular}%
}
\end{table}

\begin{table}[!h]
\centering
\caption{WebArena-Lite Hint Database Statistics by Method and Hinter Model}
\label{tab:webarena_lite_hint_stats}
\begin{tabular}{llccc}
\toprule
Hinter Method & Hinter Model & Total Entries & Unique Tasks & Avg Hints/Task \\
\midrule
\multirow{2}{*}{AutoGuide} 
 & GPT-5 & 130 & 26 & 5 \\
 & GPT-5-Mini & 130 & 26 & 5 \\
\midrule
\multirow{2}{*}{\MethodName} 
 & GPT-5 & 825 & 165 & 5 \\
 & GPT-5-Mini & 825 & 165 & 5 \\
\bottomrule
\end{tabular}
\end{table}

\subsection{Hint examples}\label{appendix:hint_examples}

\begin{figure}
    \centering
    \includegraphics[width=0.9\linewidth]{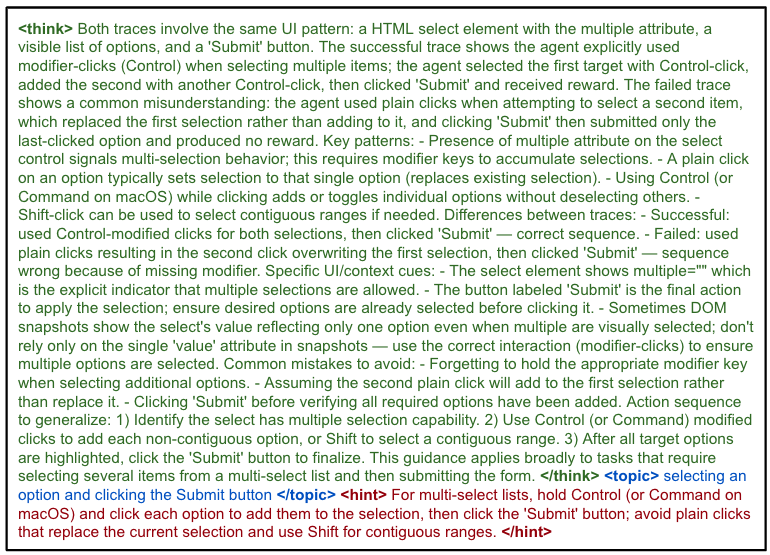}
    \caption{miniwob.click-scroll-list. \MethodName{} with gpt-5-mini as the hinter model and gpt-5-nano as the base model. The performance is improved from 0.6 to 1 on this task after applying hint.
}
\end{figure}

\begin{figure}
    \centering
    \includegraphics[width=0.9\linewidth]{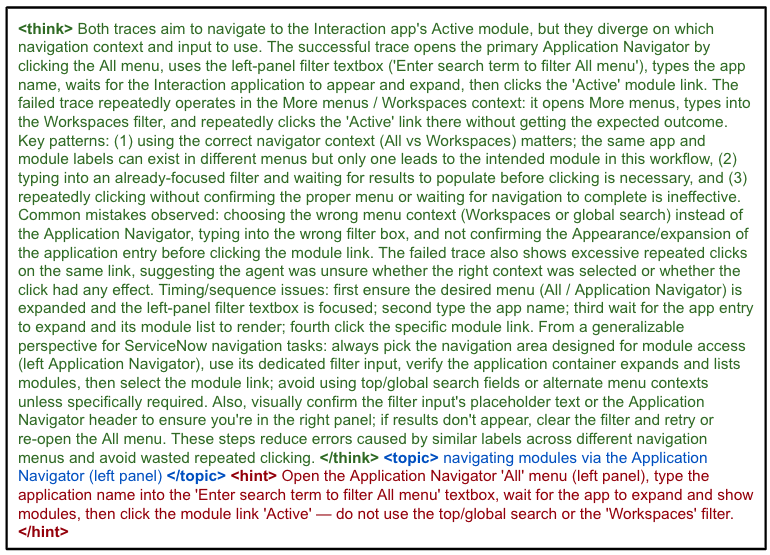}
    \caption{workarena.servicenow.all-menu. \MethodName{} with gpt-5-mini as the hinter model and gpt-5-nano as the base model. The performance is improved from 0 to 1 on this task after applying hint.
}
\end{figure}

\begin{figure}
    \label{sec:agents_leverage_hint}
    \centering
    \includegraphics[width=0.9\linewidth]{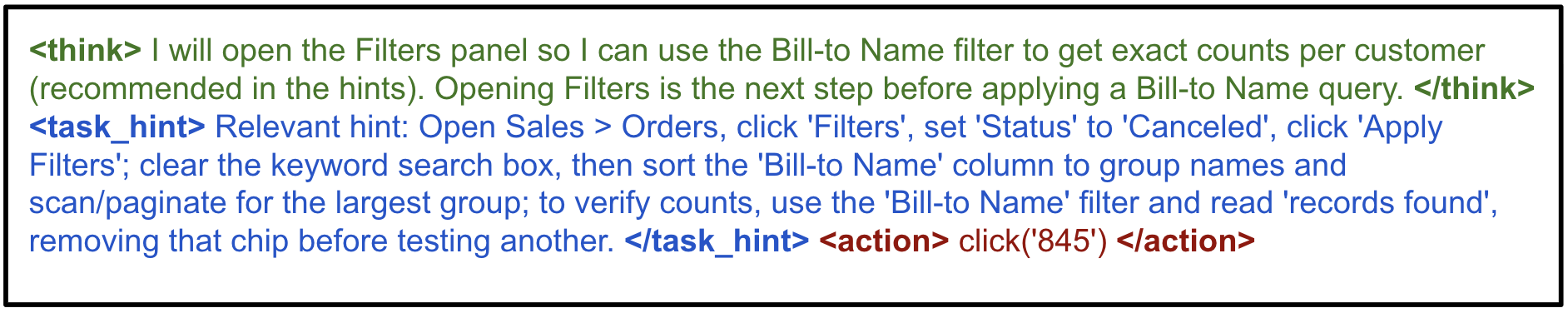}
    \caption{webarenalite.288. \MethodName{} with gpt-5 as the hinter model and gpt-5-mini as the base model. The performance is improved from 0 to 1 on this task after applying hint.
}
    \label{fig:hint_leveraged_by_agent_webarena}
\end{figure}

\newpage
\newpage
\section{Hint generation and retrieval algorithm}\label{appendix:algorithm}
\begin{algorithm}[!h]
\caption{\textbf{Hint Generation (Zoom \& Reflect)}}
\label{alg:drh_hint_generation}
\small
\begin{algorithmic}[1]
\Require Offline trajectories $\{\tau_i\}_{i=1}^N$ (with $x,z,a,r$), optional documents/instructions,\newline summarizer $\mathcal{S}$, hinter $\mathcal{H}$
\Ensure Hint database $\mathcal{D}_{\mathcal{H}}=\{(c,h)\}$
\For{\textbf{each} selection of evidence $E \in \{$single $\tau$, pair $(\tau^+,\tau^-)$, multi-trace $\{\tau^j\}_{j\in S}\}$}
  \State $c \gets \mathcal{S}(E)$ \Comment{semantic key / context used for retrieval}
  \If{\textbf{zooming}}
     \State choose set of critical steps $\mathcal{T}^\ast = \{t_1^\ast,\ldots,t_m^\ast\}$ and window $\Delta$
     \State $P \gets P_{\tau}^{\text{zoom}}=\{z,a,r\}_{1:T}\cup\bigcup_{t^\ast \in \mathcal{T}^\ast}\{x\}_{t^\ast:t^\ast+\Delta}$
  \Else
     \State $P \gets P_{\tau}^{\text{full}}=\{x,z,a,r\}_{1:T}$
  \EndIf
  \If{\textbf{contrastive}}
     \State $P \gets \text{contrastive prompt built from }(\tau^+,\tau^-)$
  \EndIf
  \State $h \gets \mathcal{H}(c, P)$ \Comment{natural–language hint linked to its source}
  \State $\mathcal{D}_{\mathcal{H}} \gets \mathcal{D}_{\mathcal{H}}\cup \{(c,h)\}$
\EndFor
\State \Return $\mathcal{D}_{\mathcal{H}}$
\end{algorithmic}
\end{algorithm}

\begin{algorithm}[!h]
\caption{\textbf{Retrieve \& Act}}
\label{alg:drh_retrieve_act}
\small
\begin{algorithmic}[1]
\Require Policy $\pi$, database $\mathcal{D}_{\mathcal{H}}=\{(c,h)\}$, retriever $\rho$, summarizer $\mathcal{S}$, goal $g$, mode $\in\{\textsc{episode},\textsc{step}\}$
\If{mode $=$ \textsc{episode}} \Comment{goal-conditioned (episode-level) retrieval}
  \State $\{h^{1},\ldots,h^{k}\}\gets \rho(g,\mathcal{D}_{\mathcal{H}})$
\EndIf
\For{$t=1,\ldots,T$}
  \State Observe $x_t$ and update $\tau_{:t}$
  \If{mode $=$ \textsc{step}} \Comment{contextual (step-level) retrieval}
     \State Define $\tau'_{:t} = (\{z,a,r\}_{1:t-1}, x_t)$
     \State $c_t \gets \mathcal{S}(\tau'_{:t})$
     \State $\{h_t^{1},\ldots,h_t^{k}\}\gets \rho(c_t,\mathcal{D}_{\mathcal{H}})$
     \State $a_t \sim \pi\!\big(x_{0:t},\{h_t^{1},\ldots,h_t^{k}\}\big)$
  \Else \Comment{\textsc{episode}}
     \State $a_t \sim \pi\!\big(x_{0:t},\{h^{1},\ldots,h^{k}\}\big)$
  \EndIf
  \State Execute $a_t$, receive $(x_{t+1}, r_t)$
\EndFor
\State \Return $\{a_t\}_{t=1}^T$
\end{algorithmic}
\end{algorithm}


\end{document}